\documentclass[10pt,twocolumn,letterpaper]{article}

\usepackage{iccv}
\usepackage{times}
\usepackage{epsfig}
\usepackage{graphicx}
\usepackage{amsmath}
\usepackage{amssymb}
\usepackage{booktabs}
\usepackage{caption}
\usepackage{multirow}
\usepackage{makecell}
\usepackage{color}
\usepackage{subcaption}
\usepackage{bbding}
\usepackage{soul}
\usepackage{float}
\usepackage[dvipsnames]{xcolor}
\usepackage{setspace}

\usepackage[breaklinks=true,bookmarks=false]{hyperref}

\hypersetup{
    colorlinks=true,    
    citecolor=green, 
    filecolor=magenta,  
    urlcolor=WildStrawberry       
}
\usepackage[accsupp]{axessibility}

\iccvfinalcopy 

\def\iccvPaperID{1748} 

\ificcvfinal\fi

\begin{document}

\title{DReg-NeRF: Deep Registration for Neural Radiance Fields}

\author{
Yu Chen \qquad\qquad\qquad Gim Hee Lee\\
Department of Computer Science, National University of Singapore\\
{\tt\small chenyu@comp.nus.edu.sg}\qquad {\tt\small gimhee.lee@nus.edu.sg} \\
}

\twocolumn[{
\renewcommand\twocolumn[1][]{#1}
\maketitle
\vspace*{-0.3in}
\centering
\captionsetup{type=figure}\includegraphics[width=0.9\textwidth]{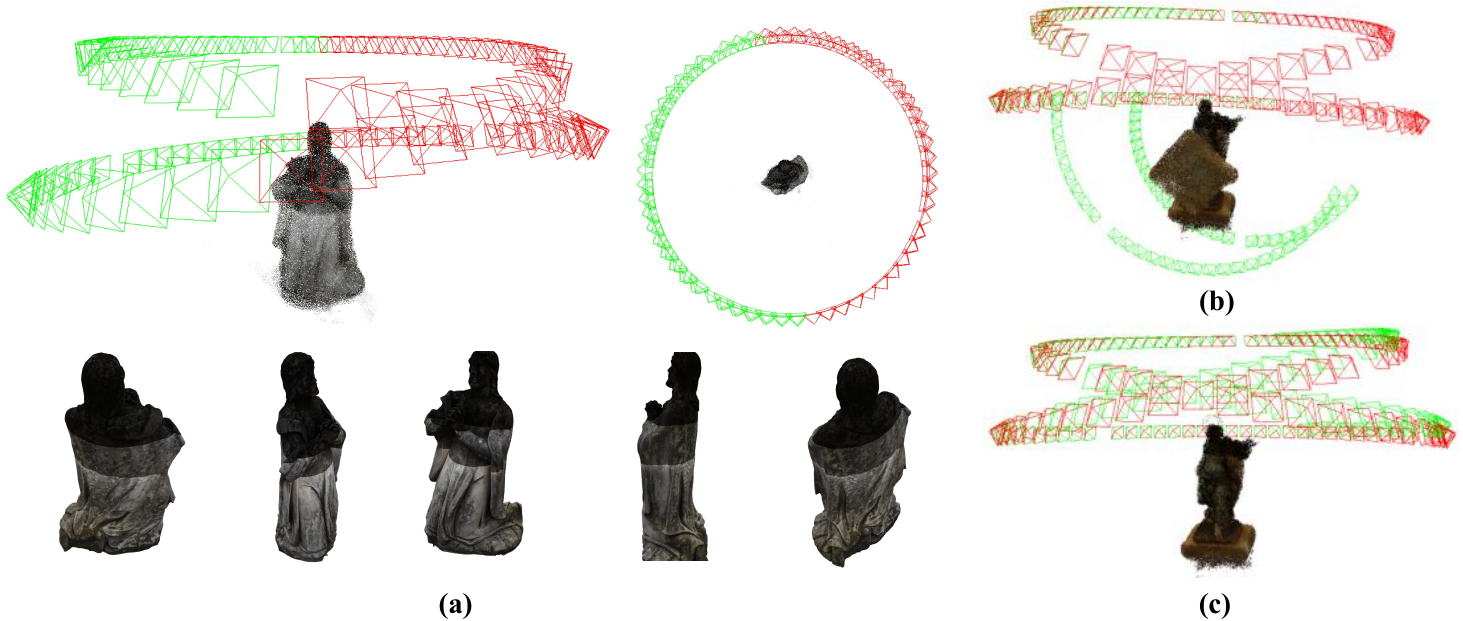}
\vspace{-2mm}
\captionof{figure}{
   Our method registers multiple NeRF blocks. (a) One of the collected objects from the 
   Objaverse~\cite{DBLP:journals/corr/abs-2212-08051} dataset. We render the
   images from a predefined camera trajectory to construct our training data. 
   (b) NeRF models are trained in different coordinate frames.
   (c) Our method aligns NeRF blocks into the same coordinate frame without accessing raw image data.
}
\label{fig:teaser}
\vspace{3mm}
}]

\maketitle
\ificcvfinal\thispagestyle{empty}\fi

\begin{abstract}
  \vspace{-4mm}
    Although Neural Radiance Fields (NeRF) is popular in the computer vision community recently, registering multiple NeRFs has yet to gain 
    much attention. 
    Unlike the existing work, NeRF2NeRF~\cite{DBLP:journals/corr/abs-2211-01600}, which is based on traditional optimization 
    methods and needs human annotated keypoints, we propose DReg-NeRF to 
    solve the NeRF registration problem on object-centric scenes without human intervention. After training NeRF models, 
    our DReg-NeRF first extracts features from the occupancy grid in NeRF. Subsequently, our DReg-NeRF utilizes a transformer architecture with 
    self-attention and cross-attention layers to learn the relations between pairwise NeRF blocks. 
    In contrast to state-of-the-art (SOTA) point cloud registration methods, the decoupled correspondences are supervised by surface fields 
    without any ground truth overlapping labels. We construct a novel view synthesis dataset with 1,700+ 3D objects obtained from Objaverse to 
    train our network. When evaluated on the test set, our proposed method beats the SOTA point cloud registration 
    methods by a large margin 
    with a mean $\text{RPE}=9.67^{\circ}$ and a mean $\text{RTE}=0.038$. 
    Our code is available at \href{https://github.com/AIBluefisher/DReg-NeRF}{https://github.com/AIBluefisher/DReg-NeRF}.
  \vspace{-0.2in}
\end{abstract}

\section{Introduction}
Scene reconstruction has many applications in the real world, for example, in augmented reality, ancient culture preservation, 3D content 
generation, \etc. Recently, rapid progress has been made in increasing the reconstruction quality using neural radiance fields (NeRF)~\cite{DBLP:conf/eccv/MildenhallSTBRN20}.
%
While previous works mostly focused on synthesizing images at the object level or unbounded scenes within a small area, 
Block-NeRF~\cite{DBLP:conf/cvpr/TancikCYPMSBK22} extends NeRF to city-scale scenes by splitting data into multiple intersected blocks.
Specifically, Block-NeRF trains multiple NeRF models in the same 
coordinate frame with the ground-truth camera poses provided by fusing multiple high-precision sensors. 
However, images can be collected without absolute pose information in some cases, \eg, when images are captured with digital cameras 
or in GPS-denied areas. In such cases, Block-NeRF cannot work since NeRF models are trained on different coordinate frames. 
Consequently, NeRF registration~\cite{DBLP:journals/corr/abs-2211-01600} is necessary for synthesizing consistent novel views from 
multiple NeRFs trained in different coordinate frames.

Point cloud registration is a classic problem in 3D computer vision, which aims at computing the relative 
transformation from the source point cloud to the target point cloud. 
However, NeRF registration is under-explored since existing works focus mostly on point cloud registration.
Unlike point clouds that are simple explicit representation, NeRF encodes scenes 
implicitly, which makes registering multiple NeRFs more challenging. NeRF2NeRF~\cite{DBLP:journals/corr/abs-2211-01600} is 
the first work that tried to solve registering NeRFs by a traditional optimization-based approach. However, it requires human-annotated keypoints 
for initialization, which limits its application in the real world where 
human annotations can be impractical. 
In view of the above-mentioned challenges, we focus on the study of the NeRF registration problem by answering the following two questions: 1) 
Can we register two or multiple NeRFs together where 
only pre-trained models are accessible?
2) 
How to register NeRFs without any human 
annotations and initializations? 

We further use the following settings in our endeavor to answer the two challenging questions on NeRF registration: 1) Images are collected 
into different blocks and no images are associated with known absolute position information. 2) Multiple NeRF blocks are trained individually 
where ground truth camera poses in each block are in their local coordinate frames. 3) Only the trained NeRF models are accessible, and 
all training images are removed and therefore not available due to plausible privacy-preserving issues or disk limitations. We emphasize that NeRF 
registration is a challenging task, and we focus more on the object-centric scenes in this paper.
See Fig.~\ref{fig:teaser} for the illustration of our task setting and dataset construction.

To solve the NeRF registration problem, we first utilize an occupancy grid along with each NeRF model to extract a voxel grid. The 
voxel grid is then fed into a 3D Feature Pyramid Network (FPN)~\cite{DBLP:conf/cvpr/LinDGHHB17} to extract features. The resulting voxel feature 
grids are further processed by a transformer module. In the transformer network, we first adopt a self-attention layer to enhance the intra-feature 
representations within each voxel feature grid. We further utilize a cross-attention layer to learn the inter-feature relations between the source feature grid and the target feature grid. Finally, we use an attention head to decode the source features and target features into correspondences and confidence scores. Unlike SOTA point clouds registration methods~\cite{DBLP:conf/cvpr/HuangGUWS21,DBLP:conf/cvpr/YewL22}, we utilize NeRF as geometric supervision and thus do not rely on 
pre-computed overlapping scores to mask correspondences outside the overlapping areas.

The main contributions of our work are:
\begin{itemize}
  \item A dataset for registering multiple NeRF blocks, which is created by rendering $1,700+$ 3D objects that are 
        downloaded from the Objaverse dataset.
  \item A novel network for registering NeRF blocks which do not rely on any human annotation and initializations.

  \item Exhaustive experiments to show the accuracy and generalization ability of our method.
\end{itemize}
To the best of our knowledge, this is the first work on registering NeRFs without \textbf{a) any initializations from keypoints or other 
registration methods} and \textbf{b) precomputed ground-truth overlapping labels}.

\section{Related Work}

\paragraph{Neural Radiance Fields.}
The differentiable volume rendering technique makes neural networks suitable for encoding scene representations. 
Vanilla NeRF~\cite{DBLP:conf/eccv/MildenhallSTBRN20} needs to take days to train each scene. NSVF~\cite{DBLP:conf/nips/LiuGLCT20} encodes 
scenes explicitly in a sparse voxel grid with online pruning, which improves the training and rendering efficiency by a large margin. 
PlenOctree~\cite{DBLP:conf/iccv/YuLT0NK21} adopted the voxel representation, and further decouple the radiance field from view directions 
by spherical harmonics. To further accelerate the network training, PlenOctree finetunes the voxel grid by directly optimizing the 
stored features without accessing any neural networks. Plenoxels~\cite{DBLP:conf/cvpr/Fridovich-KeilY22} optimizes
voxel features without any neural network for initialization. The training speed is accelerated with a specifically designed CUDA optimizer.
InstantNGP~\cite{DBLP:journals/tog/MullerESK22} encodes voxel features implicitly into a multi-resolution hash grid. 
TensoRF~\cite{DBLP:conf/eccv/ChenXGYS22} decomposes 3D voxel into low-rank vector-matrix multiplications. Its training speed is comparable 
to InstantNGP even without a CUDA implementation. 

Block-NeRF~\cite{DBLP:conf/cvpr/TancikCYPMSBK22} partition the collected images into different street blocks, each block is trained individually 
along with jointly optimizing camera poses~\cite{DBLP:conf/iccv/LinM0L21,chen2023dbarf}. The 
final images can be synthesized by fusing from multiple nearby NeRF blocks. Mega-NeRF~\cite{DBLP:conf/cvpr/TurkiRS22} also adopted the same 
divide-and-conquer strategy as BlockNeRF, but focus more on aerial images and exploiting the sparse network structures. To train their 
network, Block-NeRF obtained camera poses by fusing data from different sensors, while Mega-NeRF used Structure from Motion 
(SfM)~\cite{DBLP:conf/iccv/LindenbergerSLP21,DBLP:conf/cvpr/Chen0K21,DBLP:journals/pr/ChenSCW20,DBLP:journals/corr/abs-2301-12135} tools to recover camera poses. 
Both of them assume camera poses are in the same coordinate frames.

\vspace{-4mm}
\paragraph{Point Cloud Registration.} The Iterative Closest Point (ICP)~\cite{DBLP:journals/pami/BeslM92,DBLP:journals/ivc/ChenM92} algorithm 
has been widely applied to the industry community and research community for years. Given a rough initialization, ICP tries to align the source 
point cloud to the target point cloud. The global point cloud 
registration methods can align point clouds without initialization by extracting geometric features such as the Fast Point Feature 
Histograms (FPFH)~\cite{DBLP:conf/icra/RusuBB09}. To solve the long-time issue of the global registration method in evaluating candidate 
models within RANSAC~\cite{DBLP:journals/cacm/FischlerB81}, Zhou and Park~\cite{DBLP:conf/eccv/ZhouPK16} proposed a fast approach that does not need to evaluate the candidate 
models at each iteration. Deep point cloud registration methods also gained much attention nowadays. Deep Closest Point
~\cite{DBLP:conf/iccv/WangS19} is a learned variant of the classical ICP. Inspired by SuperGlue~\cite{DBLP:conf/cvpr/SarlinDMR20}, 
which is a deep learning method for matching 2D image correspondences, Predator~\cite{DBLP:conf/cvpr/HuangGUWS21} and 
REGTR~\cite{DBLP:conf/cvpr/YewL22} adopted the self-attention and cross-attention mechanisms from SuperGlue to learn the 
correlation for pairwise low-overlapping point clouds. The ground-truth 
overlapping scores are computed from dense point clouds and used to mask out the correspondences outside the overlapping regions. 
We also follow the previous attention mechanisms, but do not rely on the pre-computed overlapping labels.

\vspace{-4mm}
\paragraph{NeRF Registration.} NeRF2NeRF~\cite{DBLP:journals/corr/abs-2211-01600} is the first work that tries to register multiple NeRFs. 
The initial transformation is estimated 
from human-annotated keypoints, and then refined by the surface fields from NeRF. To reduce the number of useless samples, NeRF2NeRF adopts
Metropolis-Hasting sampling to maintain an active set. The whole framework is based on traditional optimization methods and needs human 
interaction. ZeroNeRF~\cite{DBLP:journals/corr/abs-2211-12544} claims it can register NeRF without overlap. However, it still requires a 
global registration method for initialization. Unlike the previous works, which rely heavily on traditional optimization methods, we try 
to register multiple NeRF blocks by learning methods without human interaction.

\section{Our Method}

\begin{figure*}[htbp]
   \centering
   \includegraphics[width=0.98\linewidth]{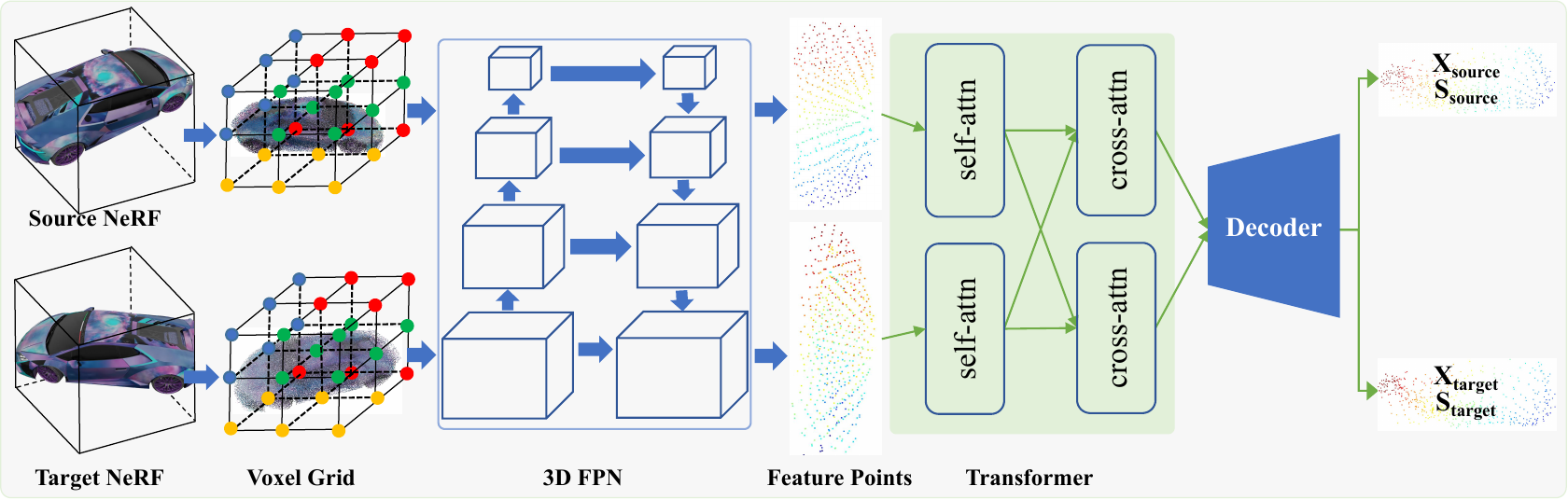}
   \vspace{-2mm}

   \caption{\textbf{Network architecture of our method}. The pipeline of our method is: 1) We first extract the pairwise 
           voxel grid $\mathbf{G} \in \mathbb{R}^{x_{\text{ref}} \times y_{\text{res}} \times z_{\text{res}} \times C}$ and a binary mask 
           $\mathbf{M} \in \mathbb{R}^{x_{\text{ref}} \times y_{\text{res}} \times z_{\text{res}} \times 1}$ from the source NeRF and 
           the target NeRF. 2) The voxel grid $\mathbf{G}$ and a binary mask $\mathbf{M}$
           are fed into the 3D feature pyramid network to extract voxel features. 3) The extracted voxel grid features are downsampled 
           to a 2-dimensional tensor $\mathcal{F} \in \mathbb{R}^{N\times C^{'}}$ by their spherical neighborhood. 4) The resulting 
           source features $\mathcal{F}_{\text{source}}$ and target features $\mathcal{F}_{\text{target}}$ are strengthened by a 
           transformer, where a self-attention layer is used to enhance the intra-contextual relations, and a cross-attention layer is 
           used to learn the inter-contextual relations. 5) Finally, we use a single-head attention layer to decode the features into 
           correspondences and their corresponding confidence scores.
           }
  \label{fig:network_architecture}
\end{figure*}

Fig.~\ref{fig:network_architecture} shows an illustration of our framework.
Our network takes 
a source NeRF model and a target NeRF model as input, and outputs correspondences in the source NeRF and target NeRF.
We first train multiple NeRF blocks in different coordinate frames. For each NeRF model, we associate it with an occupancy voxel grid, where each voxel indicates whether it is occupied or not. After training, we extract a 3D voxel grid for each NeRF model 
and then feed it into a 3D CNN backbone to extract features. Subsequently, we use a transformer with self-attention and cross-attention 
layers to learn the relations between the pairwise feature grids. We then adopt a decoder to decode the resulting features 
$\mathcal{F}_{\text{source}}, \mathcal{F}_{\text{target}}$ into correspondences 
$\{\mathbf{X}_{\text{source}}, \mathbf{X}_{\text{target}}\}$ and the corresponding confidence 
scores $\{\mathbf{S}_{\text{source}}, \mathbf{S}_{\text{target}}\}$. Finally, the relative transformation can be solved 
by the weighted Kabsch-Umeyama algorithm~\cite{DBLP:journals/pami/Umeyama91} from the correspondences.

\subsection{Background}

\paragraph{Neural Radiance Fields.} Neural Radiance Fields (NeRF) aims at rendering photo-realistic images from a new view point. For a 
3D point $\mathbf{X}$, the \textbf{density field} $\sigma_t$ of $\mathbf{X}$ is defined as the \textit{differential probability of a ray 
$\mathbf{r} = (\mathbf{o}, \mathbf{d}) $ hitting a particle}, where $\mathbf{o}$ is the camera center, $\mathbf{d}$ is the view direction.
The \textbf{transmittance} $\mathcal{T}(t)$ denotes \textit{the probability of ray without hitting any particles when traveling 
a distance $t$}, and the discrete form of $\mathcal{T}(t)$ is:
\begin{equation}
  \mathcal{T}_n = \mathcal{T}(0 \rightarrow t_n) = \exp \big( \sum_{k=1}^{n-1} - \sigma_k \delta_k \big).
\end{equation}

Given a set of points $\{\mathbf{X}_n = \mathbf{o} + t_n \mathbf{d}\ |\ n \in [0, K] \}$, NeRF predicts the view-independent volume density 
$\sigma_n$ and view-dependent radiance field by:
\begin{align}
  \label{equ:nerf_query_density_radiance}
  \sigma_n, \mathbf{e}_n = \mathbf{F} (\mathbf{X}; \mathbf{\Theta}), & \nonumber \\
  \mathbf{c}_n = \mathbf{F} (\mathbf{r}, \mathbf{e}; \mathbf{\Theta}), &
\end{align}
where $\mathbf{e}_n$ is an embedding vector and $\Theta$ denotes the network parameters. The final color of an image pixel can be rendered by:
\begin{equation}
  \label{equ:vol_rendering}
  \mathbf{C} (t_{N+1}) = \sum_{n=1}^N \mathcal{T}_n \cdot \big( 1 - \exp (-\sigma_n \delta_n) \big) \cdot \mathbf{c}_n.
\end{equation}
We recommend interested readers to refer to~\cite{DBLP:conf/eccv/MildenhallSTBRN20,DBLP:journals/corr/abs-2209-02417} for the detailed 
derivation.

\subsection{Querying Radiance Fields from NeRF}
\label{subsec:query_rf}

To extract features for the latter learning modules, we first construct a 3D volume from NeRF. Specifically, we 
assume each NeRF is trained within a bounding box with a 3D voxel grid of resolution 
$[x_{\text{ref}}, y_{\text{res}}, z_{\text{res}}]$. We then obtain the point locations $\{\mathbf{X}\}$ of voxel centers. 
We also obtain a binary occupancy mask $\mathbf{M}_{\text{occ}}$ from the occupancy grid, where voxels 
that are not occupied denote empty space and thus are ignored. The occupancy grid is the acceleration structure used in 
InstantNGP~\cite{DBLP:journals/tog/MullerESK22} to skip empty space in NeRF training. Each grid has a resolution of $128^3$ 
that is centered around $(0,0,0)$ and stores occupancy as a single bit. During ray marching, a sample point is skipped if 
the bit of a grid cell is low.
%
We can query the 
density fields $\{\sigma\}$ and radiance fields $\{\mathbf{c}\}$ using 
Eq.~\eqref{equ:nerf_query_density_radiance} with the coordinates $\{\mathbf{X}\}$. Since density fields can be noisy, we further 
obtain a density mask $\mathbf{M}_{\text{df}} = \sigma > \sigma_t$ by setting a threshold $\sigma_t$ (we use $\sigma_t=0.7)$. 
We then obtain our mask as $\mathbf{M} = \mathbf{M}_{\text{occ}} \cap \mathbf{M}_{\text{df}}$.

One issue with radiance fields is that they are view-dependent. However, the training views are different for each NeRF block.
Suppose the NeRF is trained by $N$ views with camera poses 
$\mathbf{P} = [\mathbf{R} \mid \mathbf{t}]$ which project points from camera frame to world frame. To obtain the radiance fields, we form $N$ viewing rays 
$\mathbf{r} = \mathbf{o} + t \mathbf{d}$, where $\mathbf{d} = \frac{\mathbf{X} - \mathbf{t}}{\|\mathbf{X} - \mathbf{t}\|}$ for 
each query point and average the queried radiance fields over all rays. The final color $\mathbf{C}$ can be obtained by 
volume rendering of Eq.~\eqref{equ:vol_rendering}. Furthermore, we compute the alpha compositing value by 
$\alpha = 1 - \exp (-\sigma \delta)$, where $\delta$ is a chosen small value.
We associate 
each point in the voxel grid with 
$[\mathbf{X}, \mathbf{C}, \alpha]$ and 
feed the voxel grid $\mathbf{G} \in \mathbb{R}^{x_{\text{ref}} \times y_{\text{res}} \times z_{\text{res}} \times C}$ and a mask 
$\mathbf{M} \in \mathbb{R}^{x_{\text{ref}} \times y_{\text{res}} \times z_{\text{res}} \times 1}$  
into a 3D CNN to extract backbone features, where $C=7$ with $3$ channels from the point coordinate, three from the color channels, and 
one from the scalar $\alpha$.

\subsection{Feature Extraction}

Given the voxel grid, we adopt the feature pyramid network~\cite{DBLP:conf/cvpr/LinDGHHB17} to extract features. We use 
ResNet~\cite{DBLP:conf/cvpr/HeZRS16} as the feature backbone, where all the 2D modules are replaced by their corresponding 3D parts. 
The feature pyramid network enables the learning of high-level semantic features at a multi-scale and thus is suitable to extract the voxel 
grid features in our task settings. We note the difference from the original feature pyramid network which utilizes different scale features 
for object detection, we adopt only the features output from the last layer. Since the dimension of the feature grid from the last 
layer can be different from the original voxel grid, we rescale it to the size of the original voxel grid $\mathbf{G}$ and resulted in 
a voxel feature grid $\mathbf{G}_f \in \mathbb{R}^{x_{\text{ref}} \times y_{\text{res}} \times z_{\text{res}} \times C^{'}}$.

We cannot use $\mathbf{G}_f$ as the input to our transformer since the voxel feature grid can contain too many points to enable the transformer to run on a single GPU. 
To solve this issue, we iteratively downsample $\mathbf{G}_f$ by the spherical neighborhood~\cite{DBLP:conf/3dim/ThomasGDM18} as done in 
KPConv~\cite{DBLP:conf/iccv/ThomasQDMGG19} to obtain the downsampled voxel points $\hat{\mathbf{X}}$. The downsample iteration is 
terminated when the total number of occupied voxels in the current sampled voxel feature grid is less than $1.5$K. Lastly, we reshape 
the occupied voxel features into a tensor $\mathbf{\mathcal{F}} \in \mathbb{R}^{N \times C^{'}}$. The weights of the feature extraction 
network are shared across the source 
and the target NeRFs. 
We denote the downsampled voxel points and extracted features of the source and target NeRFs as:
$[\hat{\mathbf{X}}_{\text{source}}, \mathcal{F}_{\text{source}}]$ and $[\hat{\mathbf{X}}_{\text{target}}, \mathcal{F}_{\text{target}}]$, respectively.

\subsection{Transformer}
We then feed the resulting feature $\mathcal{F}_{\text{source}}$ and $\mathcal{F}_{\text{target}}$ into a $L$-layer transformer 
with self-attention and cross-attention layers. We follow the same intuition of Predator~\cite{DBLP:conf/cvpr/HuangGUWS21} and 
REGTR~\cite{DBLP:conf/cvpr/YewL22}, where a self-attention layer is applied to both the 
source feature $\mathcal{F}_{\text{source}}$ and the target feature $\mathcal{F}_{\text{target}}$ to enhance the intra-contextual 
relations, and a cross-attention layer is applied to both $\mathcal{F}_{\text{source}}$ and $\mathcal{F}_{\text{target}}$ to learn the 
inter-contextual relations. We follow the classical transformer 
architecture~\cite{DBLP:conf/nips/VaswaniSPUJGKP17,DBLP:journals/access/LiWZ22a,DBLP:conf/cvpr/YewL22} with input to be voxel features.

\vspace{-3mm}
\paragraph{Multi-Head Attention.} The multi-head attention operation in each layer is defined as:
\begin{align}
  \text{MH}(\mathbf{Q}, \mathbf{K}, \mathbf{V}) = \text{concat} (\text{head}_1, \cdots, \text{head}_h)\mathbf{W}^O,
\end{align}
where $\text{head}_i=\text{Attention} (\mathbf{Q} \mathbf{W}^Q_i, \mathbf{K} \mathbf{W}_i^K, \mathbf{V} \mathbf{W}_i^V)$. The 
attention function is adopted as the scaled dot-product:
\begin{equation}
  \text{Attention} (\mathbf{Q}, \mathbf{K}, \mathbf{V}) = \text{softmax} (\frac{\mathbf{Q} \mathbf{K}^{\top}}{\sqrt{d_k}}) \mathbf{V}.
\end{equation}

In the self-attention layers, $\mathbf{Q}=\mathbf{K}=\mathbf{V}$ represents the same feature tensor in each block. In the cross-attention 
layers, the keys, and the values are the feature tensors from the other block. The self-attention mechanism enables the network to learn the 
relationship inside the same feature points, while the cross-attention mechanism enables the communication of the different feature points.

\vspace{-3mm}
\paragraph{Decoder.} 
After encoding features by transformer, we further adopt a single-head attention layer to predict the corresponding 
point locations $\tilde{\mathbf{X}}_{\text{source}}$ and confidence scores $\tilde{\mathbf{S}}_{\text{source}}$ of the 
source voxel points $\hat{\mathbf{X}}_{\text{source}}$ in the target NeRF's coordinate frame. Similarly, we also predict the corresponding point locations $\tilde{\mathbf{X}}_{\text{target}}$ and confidence scores $\tilde{\mathbf{S}}_{\text{target}}$ of the target voxel points $\hat{\mathbf{X}}_{\text{target}}$ in the source NeRF's coordinate frame. Finally, we utilize the predicted correspondences to compute the relative rigid transformation. The confidence scores are used as weights that mask out the irrelevant 
correspondences and can be interpreted as how likely the predicted points from $\tilde{\mathbf{X}}_{\text{source}}$ and 
$\tilde{\mathbf{X}}_{\text{target}}$ are correspondences and are visible in the source NeRF and the target NeRF. 

\subsection{Training Loss}
\paragraph{Surface Field Supervision.}
To train the network, we encourage the predicted correspondences to have the same features which are invariant in the corresponding NeRF model.
The na\"ive way is to adopt density fields as supervision instead of radiance fields since density fields are invariant to view points. 
However, density fields can be very noisy. We thus utilize the surface fields~\cite{DBLP:journals/corr/abs-2211-01600} as supervision. 
The surface field is defined as the differential probability of the ray hitting a surface at $\mathbf{X}_n$ given by:
\begin{equation}
  \mathcal{S}(t) = \mathcal{T}(t) \cdot (1 - \exp(-2 \sigma \delta)).
\end{equation}
\vspace{-5mm}
\paragraph{Proof:}
  The differential probability of a ray hitting a surface at point $\mathbf{X}$ is the product of the probability of a ray 
  traveling over $[0, t_n)$ without hitting any particle before $t_n$ times the differential probability of the ray hitting exactly at 
  point $\mathbf{X}(t_n)$. The surface field can then be written as:
  \begin{equation}
    S(t) = \int_{t-\delta}^{t+\delta} \mathcal{T}(s) \cdot \sigma(s) \ ds
  \end{equation}
  We derive the exact form of $S(t)$ as follow:
  \begin{align}
    S(t) &= \int_{t-\delta}^{t+\delta} \mathcal{T}(s) \cdot \sigma(s) \ ds \nonumber \\
         &= \int_{t-\delta}^{t+\delta} \mathcal{T}(0 \rightarrow t-\delta) \cdot \mathcal{T} (t-\delta \rightarrow s) \cdot \sigma(s) \ ds \nonumber \\
         &= \mathcal{T}(t-\delta) \cdot \int_{t-\delta}^{t+\delta} \mathcal{T}(t-\delta \rightarrow s) \cdot \sigma(s)\ ds \nonumber \\
         &= \mathcal{T}(t-\delta) \cdot \sigma_t \cdot \int_{t-\delta}^{t+\delta} \mathcal{T} (t-\delta \rightarrow s) \ ds \nonumber \\
         &= \mathcal{T}(t-\delta) \cdot \sigma_t \cdot \int_{t-\delta}^{t+\delta} \Big(\exp \big(-\int_{t-\delta}^s \sigma (\mu) \ d\mu \big) \Big) \ ds \nonumber \\
         &= \mathcal{T}(t-\delta) \cdot \sigma_t \cdot \int_{t-\delta}^{t+\delta} \Big(\exp \big( -\sigma_t (s-t+\sigma)\big) \Big) \ ds \nonumber \\
         &= \mathcal{T}(t-\delta) \cdot \sigma_t \cdot (-\frac{1}{\sigma_t}) \cdot \exp \big(-\sigma_t (s-t+\delta) \big) \big|_{t-\delta}^{t+\delta} \nonumber \\
         &= \mathcal{T}(t-\delta) \cdot \big( 1 - \exp (-2\sigma_t \cdot \delta) \big).
  \end{align}
The second term holds since transmittance is multiplicative (\cf Eq.(18) of~\cite{DBLP:journals/corr/abs-2209-02417}). The 4th term holds 
since we can assume the density $\sigma_t$ is a constant within a small region $[t-\delta, t+\delta]$. 
To produce the view-independent field, we first form N viewing rays for each point. Subsequently, we take
the maximum value of all rays as the density field value instead of averaging the value (as done in Sec.~\ref{subsec:query_rf}) for the
radiance field. Finally, we obtain the surface field mask $\mathbf{M}_{\text{sf}}$ by checking $S(t) > \eta$ (we use $\eta$ = 0.5). 
We then update our mask by $\mathbf{M}=\mathbf{M}_{\text{occ}}\ \cap \ \mathbf{M}_{\text{df}}\ \cap \ \mathbf{M}_{\text{sf}}$.

\vspace{-3mm}
\paragraph{Confidence Loss.}
We adopt the cross-entropy loss to supervise the confidence score, with the surface fields 
$\mathbf{\hat{S}}_{\text{source}}=S(\hat{\mathbf{X}}_{\text{source}}), \mathbf{\hat{S}}_{\text{target}}=S(\hat{\mathbf{X}}_{\text{target}})$ 
queried from NeRF as the ground truth label:
\begin{equation}
  \mathcal{L}_{\text{conf}} = \text{BCE}(\mathbf{\hat{S}}_{\text{source}}, \tilde{\mathbf{S}}_{\text{source}}) + 
                              \text{BCE}(\mathbf{\hat{S}}_{\text{target}}, \tilde{\mathbf{S}}_{\text{target}}).
\end{equation}
\vspace{-5mm}
\paragraph{Surface Field Loss.}
In addition to the confidence loss, we also encourage the predicted correspondences to have consistent surface field values:
\begin{equation}
\begin{small}
  \mathcal{L}_{\text{sf}} = \frac{1}{N} \| 
                                    \mathbf{S} ([\hat{\mathbf{X}}_{\text{source}}, \hat{\mathbf{X}}_{\text{target}}]) -
                                    \mathbf{S} ([\tilde{\mathbf{X}}_{\text{source}}, \tilde{\mathbf{X}}_{\text{target}}])   
                                  \|_1,
\end{small}
\end{equation}
where $N$ is the total number of the concatenated source voxel points and target voxel points.
\vspace{-3mm}
\paragraph{Correspondence Loss.}
We further use a correspondence loss to constrain the predicted locations of correspondences:
\begin{equation}
  \mathcal{L}_{\text{corr}} = \sum \rho (\mathbf{S} \| \mathcal{T}^{\star} (\mathbf{x}_i) - \mathbf{y}_i \|; \eta, \gamma),
\end{equation}
where $\mathcal{T}^{\star}$ is the ground truth relative transformation between the source NeRF and target NeRF, 
$\rho(\cdot; \eta, \gamma)$ is the adaptive robust loss function~\cite{DBLP:conf/cvpr/Barron19} and $\{\eta, \gamma\}$ are respectively a 
smooth interpolation value and a scale parameter, which are hyperparameters used to control the shape of the robust function. We use 
$\eta=1.0, \gamma=0.5$ in our experiments.

Moreover, we adopt the feature loss~\cite{DBLP:journals/corr/abs-1807-03748,DBLP:conf/cvpr/YewL22} to leverage the geometric 
properties when computing the correspondences. Our final loss is therefore defined as:
\begin{equation}
  \mathcal{L}_{\text{final}} = \mathcal{L}_{\text{conf}} + \lambda_1 \mathcal{L}_{\text{sf}} + 
                               \lambda_2 \mathcal{L}_{\text{corr}} + \lambda_3 \mathcal{L}_{\text{feat}},
\end{equation}
where $\lambda_1,\lambda_2,\lambda_3$ are the weights associated with the corresponding loss functions. In our experiments, we use 
$\lambda_1=1.0, \lambda_2=0.1, \lambda_3=1.0$.

\section{Experiments}
\label{sec:experiments}

\begin{figure*}[ht]
  \centering

  \includegraphics[width=0.99\linewidth]{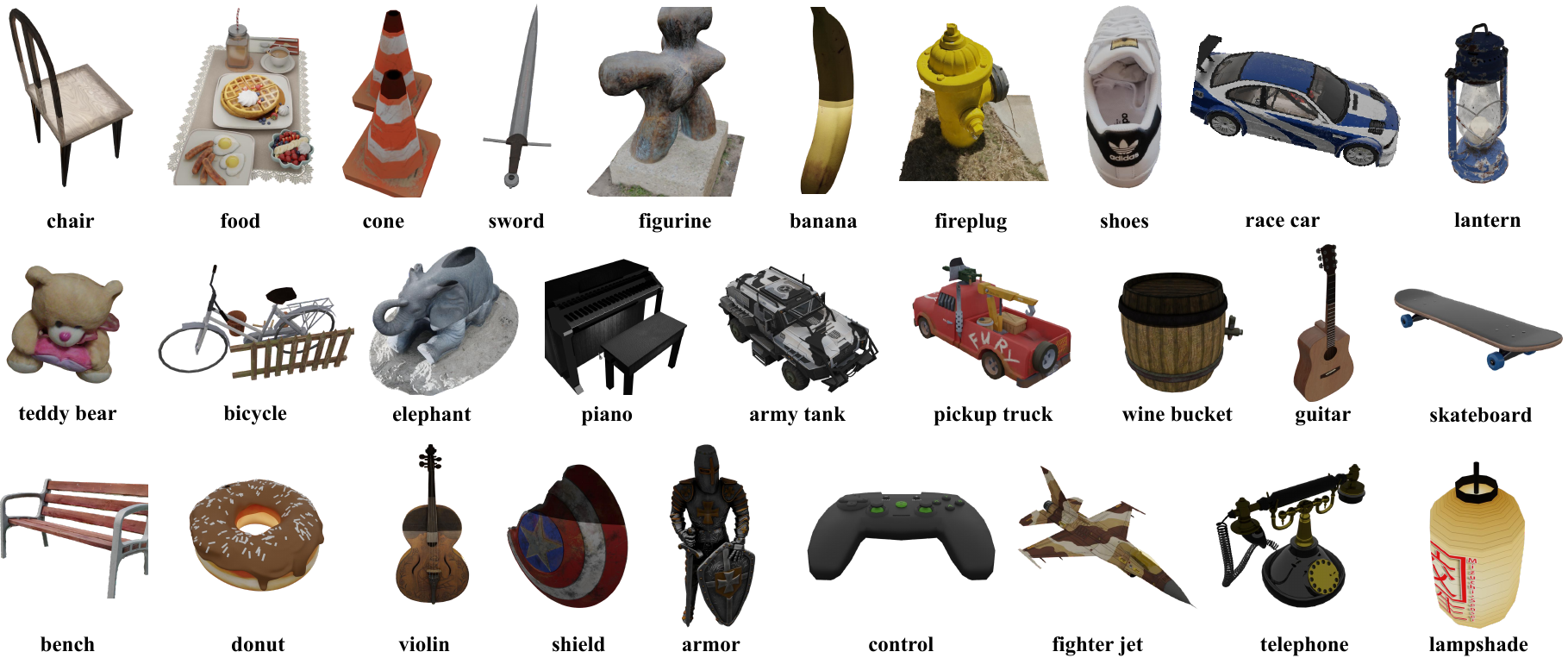}

  \caption{\textbf{An overview of our training data}. The training images are rendered from 3D objects collected from the 
           Objaverse dataset. We randomly picked $\sim 30$ classes, on which each class contains 40--80 objects.}

  \label{fig:training_data_overview}
\end{figure*}

\paragraph{Datasets.}
We aim at registering multiple NeRFs. Due to the lack of a suitable dataset for our task, we downloaded 
the 3D mesh models of $1,700+$ objects from Objaverse~\cite{DBLP:journals/corr/abs-2212-08051} to construct our dataset. 
Objaverse~\cite{DBLP:journals/corr/abs-2212-08051} 
is a massive dataset that contains $800K+$ annotated 3D Objects. It is created for the text-to-3D task. We 
utilize it to construct our dataset for NeRF registration. Specifically, we randomly selected $30+$ categories and each category 
contains $40-80$ objects. As Objaverse contains only 3D objects, we render 120 images for each object where the distribution of camera 
poses can be seen from Fig.~\ref{fig:teaser} (a). We then split the images into 2 blocks by KMeans. We also add a randomly generated transformation to the original camera poses after splitting data into separate 
blocks, such that NeRF blocks are trained in different 
coordinate frames. Each NeRF block is trained in $10K$ iterations. See Fig.~\ref{fig:training_data_overview} for an overview of our 
selected training data. 
The NeRF models trained on all objects are used to train our NeRF registration neural network, and we randomly select 44 objects that are not seen during training for the test.
\vspace{-5mm}
\paragraph{Implementations.} 
We render images for all $1,700+$ objects in a computer with an Intel i7 CPU and an NVIDIA GTX 4090 GPU. We run 8 processes concurrently 
for downloading the mesh models and the job is finished within a week. We use an occupancy grid with a resolution of $128\times 128 \times 128$ associated 
InstantNGP~\cite{DBLP:journals/tog/MullerESK22} as our NeRF representation. To train NeRF, we sample 1024 points for each ray. We use 
Adam~\cite{DBLP:journals/corr/KingmaB14} as the NeRF optimizer. We set the initial learning rate to $1e-2$ and decay it at step $5K, 7.5K, 9K$ with 
a multiplicative factor being $0.33$.
Except for storing the NeRF models, we also store the camera 
poses and intrinsics as metadata. For our NeRF registration network, we use AdamW~\cite{DBLP:conf/iclr/LoshchilovH19} as the optimizer with 
weight decay $1e-4$. The learning rate is set to $1e-4$ and halved every $34K$ iteration. The batch size is $1$. Our network is trained 
for $60$ epochs, which took about 48 hours to finish. For our transformer, we use $L=6$ layers and $h=8$ heads.
\begin{table*}[!h]
  \centering
  \resizebox{0.98\textwidth}{!}{
    \begin{tabular}{c | l r r r r r r r r r r r r}
      \toprule

        &  & 
        \multicolumn{1}{l}{\textbf{Food 5648}} &
        \multicolumn{1}{l}{\textbf{Chair 4b05}} &  
        \multicolumn{1}{l}{\textbf{Chair 4659}} &
        \multicolumn{1}{l}{\textbf{Chair 3f2d}} &
        \multicolumn{1}{l}{\textbf{Cone 37b5}} &
        \multicolumn{1}{l}{\textbf{Figurine 260d}} & 
        \multicolumn{1}{l}{\textbf{Figurine 0a5b}} &
        \multicolumn{1}{l}{\textbf{Figurine 09f0}} &
        \multicolumn{1}{l}{\textbf{Banana 3a07}} &
        \multicolumn{1}{l}{\textbf{Banana 2373}} &
        \multicolumn{1}{l}{\textbf{Banana 0a07}} 
      \\
      
      \midrule

      \multirow{4}{*}{$\Delta \mathbf{R}$} 
      & FGR~\cite{DBLP:conf/eccv/ZhouPK16} & 178.34 &  50.50 &  28.54 &  81.31 & 104.52 &  89.13 
                                           &  26.35 & 138.00 &  12.17 &   6.92 &   2.86 
                                           \\
      & REGTR~\cite{DBLP:conf/cvpr/YewL22} & 169.07 & 150.38 &  92.80 &  98.67 &  62.50 & 111.80 
                                           & 106.12 & 176.48 & 136.02 & 178.36 & 173.96 
                                           \\
      & $\text{Ours}_{\text{df}}$ & 77.48 & 160.13 & 157.21 & 22.91 & 108.09 & 121.32 
                                  & 10.53 & 95.89 & 95.43 & \textbf{3.49} & 6.96  \\
      & Ours             & \textbf{6.01} &  \textbf{6.53} & \textbf{17.74} & \textbf{18.88} & \textbf{18.79} & \textbf{2.11} 
                         & \textbf{7.62} & \textbf{8.25} & \textbf{15.55} & 10.95 & \textbf{1.36} 
                         \\
      
      \hline
      
      \multirow{4}{*}{$\Delta \mathbf{t}$} 
      & FGR~\cite{DBLP:conf/eccv/ZhouPK16} & 17.44 &  \textbf{2.27} &  \textbf{7.10} &  8.65 &  30.49 & 19.25 
                                           & 10.93 & 35.22 &  8.50 & 1.53  &  1.36  
                                           \\
      & REGTR~\cite{DBLP:conf/cvpr/YewL22} & 30.72 & 15.41 & 24.97 & 60.53 &  84.20 & 62.07 
                                           & 35.48 & 42.10 & 10.75 & 50.40 & 13.17 
                                           \\
      & $\text{Ours}_{\text{df}}$ & 15.52 & 7.32 & 11.72 & \textbf{2.29} & 21.70 & 33.61 
                                  & \textbf{1.95} & 21.40 & 13.14 & 4.28 & \textbf{0.50} \\
      & Ours             &  \textbf{1.78} & 4.13 & 8.74 & 5.07 & \textbf{3.06} & \textbf{3.54} 
                         & 10.68 & \textbf{3.18} & \textbf{0.46}  & \textbf{1.00} & 1.22 
                         \\

      \bottomrule
    \end{tabular}
  }
  
  \resizebox{0.98\textwidth}{!}{
    \begin{tabular}{c | l r r r r r r r r r r r r}
      \toprule

        &  & 
      \multicolumn{1}{l}{\textbf{Fireplug 06d5}} &
      \multicolumn{1}{l}{\textbf{Fireplug 0063}} &
      \multicolumn{1}{l}{\textbf{Fireplug 0152}} &  
      \multicolumn{1}{l}{\textbf{Shoe 18c3}} &
      \multicolumn{1}{l}{\textbf{Shoe 1627}} &
      \multicolumn{1}{l}{\textbf{Shoe 0bf9}} &
      \multicolumn{1}{l}{\textbf{Shoe 022c}} & 
      \multicolumn{1}{l}{\textbf{Teddy 1b47}} &
      \multicolumn{1}{l}{\textbf{Elephant 183a}} &
      \multicolumn{1}{l}{\textbf{Elephant 1608}} &
      \multicolumn{1}{l}{\textbf{Elephant 1a39}} 
      \\
      
      \midrule

      \multirow{4}{*}{$\Delta \mathbf{R}$} 
      & FGR~\cite{DBLP:conf/eccv/ZhouPK16} & \textbf{6.19} & 20.32 &  7.50 & 10.23 & 178.14 & 71.55 
                                           & 50.28 & 8.05 & 7.65 &  21.37 & 30.97 
                                           \\
      & REGTR~\cite{DBLP:conf/cvpr/YewL22}  & 156.92 & 99.60 & \textbf{4.04} & \textbf{2.55} & 175.21 &  97.92 
                                            &  154.91 & 149.17 & 177.15 & 172.28 & 102.62 
                                            \\
      & $\text{Ours}_{\text{df}}$ & 156.17 & 45.76 & 12.34 & 14.69 & 131.66 & 158.66 
                                  & 6.84 & \textbf{6.32} & \textbf{6.97} & \textbf{3.92} & 126.94
                                  \\
      & Ours             & 7.96 & \textbf{17.43} &  4.86 & 6.06 & \textbf{12.95} & \textbf{6.48} 
                         & \textbf{2.93} & 11.44 & 8.00 & 11.13 & \textbf{13.84} 
                         \\
      
      \hline
      
      \multirow{4}{*}{$\Delta \mathbf{t}$} 
      & FGR~\cite{DBLP:conf/eccv/ZhouPK16} & 5.83 &  \textbf{0.83} &  1.17 &  \textbf{0.04} & \textbf{4.99} & 8.82 
                                           &  35.47 &  1.11 & 4.51 & 14.08 & 11.03 
                                           \\
      & REGTR~\cite{DBLP:conf/cvpr/YewL22} & 68.71 & 38.74 & 2.13 & 3.53 & 43.40 & 61.37 
                                           & 102.00 & 42.84 & 52.26 & 66.15 & 34.54 
                                           \\
      & $\text{Ours}_{\text{df}}$ & 10.54 & 5.32 & 2.60 & 4.66 & 28.63 & 24.82 
                                  & 4.40 & 2.20 & \textbf{4.26} & \textbf{1.40} & 33.57  
                                  \\
      & Ours             & \textbf{1.58} & 5.08 & \textbf{0.96} & 2.08 & 12.80  & \textbf{1.81}  
                         & \textbf{0.65} & \textbf{1.06} & 8.97 & 6.17 & \textbf{7.80} 
                         \\

      \bottomrule
    \end{tabular}
  }

  \caption{Quantitative results (first part) of registration on the Objaverse dataset. 
           $\Delta \mathbf{R}$ denotes the relative rotation errors 
           in degree, $\Delta \mathbf{t}$ denotes the relative translation errors multiplied by 1e2 with unknown scales. 
           FGR~\cite{DBLP:conf/eccv/ZhouPK16} denotes the fast global matching method, $\text{Ours}_{\text{df}}$ denotes our 
           method with surface fields replaced by density fields.} 

  \label{table:quantitive_objaverse_registration_part12}
\end{table*}

\begin{table*}[!h]
   \centering
   
   \resizebox{0.98\textwidth}{!}{
     \begin{tabular}{c | l r r r r r r r r r r r r}
       \toprule
 
         &  & 
         \multicolumn{1}{l}{\textbf{Piano 0e0d}} &
         \multicolumn{1}{l}{\textbf{Piano 0a6e}} &
         \multicolumn{1}{l}{\textbf{Truck 1431}} &
         \multicolumn{1}{l}{\textbf{Guitar 15b4}} &  
         \multicolumn{1}{l}{\textbf{Guitar 14f8}} &
         \multicolumn{1}{l}{\textbf{Guitar 0ceb}} &
         \multicolumn{1}{l}{\textbf{Guitar 0aa0}} &
         \multicolumn{1}{l}{\textbf{Lantern 0231}} & 
         \multicolumn{1}{l}{\textbf{Lamp 0230}} &
         \multicolumn{1}{l}{\textbf{Bench 0b05}} &
         \multicolumn{1}{l}{\textbf{Shield 22a7}} 
       \\
       
       \midrule
 
       \multirow{4}{*}{$\Delta \mathbf{R}$} 
       & FGR~\cite{DBLP:conf/eccv/ZhouPK16} & 23.09 & 77.63 & \textbf{7.46} & \textbf{7.80} & 5.25 & 13.07  
                                            & 39.94 & 130.36 & 17.44 & 19.51 & 170.27   
                                            \\
       & REGTR~\cite{DBLP:conf/cvpr/YewL22} & 30.54 & 117.90 & 178.49 & 5.18 & 29.47 & 103.84
                                            & \textbf{5.95} & 139.32 & 160.45 & 122.12 & 157.38
                                            \\
       & $\text{Ours}_{\text{df}}$ & 160.71 & 168.79 & 117.07 & 11.43 & 164.87 & 177.62 
                                   & 7.96 & \textbf{7.76} & 173.09 & 179.16 & 178.00 \\
       & Ours             & \textbf{16.30} & \textbf{13.51} & 16.68 & 12.60 & \textbf{3.43} & \textbf{1.08}
                          & 9.53 & 9.17 & \textbf{16.44} & \textbf{12.98} & \textbf{8.21}
                          \\
       
       \hline
       
       \multirow{4}{*}{$\Delta \mathbf{t}$} 
       & FGR~\cite{DBLP:conf/eccv/ZhouPK16} & 7.43 & 14.50 & 5.95 & \textbf{2.86} & \textbf{3.46} & \textbf{1.83}
                                            & 8.42 & 9.06 & \textbf{0.69} & 12.52 & 15.57
                                            \\
       & REGTR~\cite{DBLP:conf/cvpr/YewL22} & 44.24 & 65.99 & 50.63 & 15.18 & 18.41 & 89.20
                                            & 9.91 & 57.25 & 64.44 & 31.97 & 44.29
                                            \\
       & $\text{Ours}_{\text{df}}$ & 22.86 & 26.18 & 24.83 & 10.27 & 8.50 & 43.21 
                                   & 4.09 & 3.35 & 26.77 & 28.59 & 34.71 \\
       & Ours             & \textbf{4.80} & \textbf{12.54} & \textbf{0.04} & 5.72 & 5.03 & 3.01
                          & \textbf{1.20} & \textbf{3.29} & 1.31 & \textbf{1.68} & \textbf{11.33}
                          \\
 
       \bottomrule
     \end{tabular}
   }
   
   \resizebox{0.98\textwidth}{!}{
     \begin{tabular}{c | l r r r r r r r r r r r r}
       \toprule
 
         &  & 
       \multicolumn{1}{l}{\textbf{Shield 1973}} &
       \multicolumn{1}{l}{\textbf{Shield 14a6}} &
       \multicolumn{1}{l}{\textbf{Shield 00ad}} &
       \multicolumn{1}{l}{\textbf{Controller 0866}} &
       \multicolumn{1}{l}{\textbf{Fighter Jet 16c6}} &  
       \multicolumn{1}{l}{\textbf{Fighter Jet 089f}} &
       \multicolumn{1}{l}{\textbf{Fighter Jet 0000}} &
       \multicolumn{1}{l}{\textbf{Telephone 1a8c}} &
       \multicolumn{1}{l}{\textbf{Telephone 0354}} & 
       \multicolumn{1}{l}{\textbf{Lampshade ab66}} &
       \multicolumn{1}{l}{\textbf{Skateboard 10c7}} 
       \\
       
       \midrule
 
       \multirow{4}{*}{$\Delta \mathbf{R}$} 
       & FGR~\cite{DBLP:conf/eccv/ZhouPK16} & 130.83 & 178.99 & 7.06 & 164.01 & 11.91 & 40.21
                                            & 39.86 & 150.10 & 19.76 & 147.50 & 176.90
                                            \\
       & REGTR~\cite{DBLP:conf/cvpr/YewL22}  & 138.94 & 169.78 & 14.76 & 102.05 & 154.74 & 150.64
                                             & 178.35 & 144.47 & \textbf{1.13} & 148.44 & 3.88
                                             \\
       & $\text{Ours}_{\text{df}}$ & 178.93 & \textbf{4.92} & 7.78 & 179.13 & 9.75 & 23.97 
                                   & 178.68 & 132.49 & 16.59 & 5.79 & 179.97 \\
       & Ours             & \textbf{12.94} & 12.26 & \textbf{2.29} & \textbf{4.03} & \textbf{6.88} & \textbf{10.53}
                          & \textbf{6.46} & \textbf{15.60} & 9.01 & \textbf{5.67} & \textbf{1.92}
                          \\
       
       \hline
       
       \multirow{4}{*}{$\Delta \mathbf{t}$} 
       & FGR~\cite{DBLP:conf/eccv/ZhouPK16} & 49.44 & 55.62 & \textbf{0.22} & 8.97 & 6.90 & 11.73
                                            & 3.44 & 57.53 & 18.51 & 67.08 & 19.03
                                            \\
       & REGTR~\cite{DBLP:conf/cvpr/YewL22} & 72.01 & 48.56 & 6.57 & 65.87 & 52.97 & 82.10
                                            & 28.79 & 51.09 & \textbf{0.91} & 54.45 & 5.14
                                            \\
       & $\text{Ours}_{\text{df}}$ & 59.08 & \textbf{0.66} & 1.58 & 23.68 & 2.64 & 13.98 & 19.28 & 63.76 & 15.65 & 4.81 & 12.20 \\
       & Ours             & \textbf{2.79} & 4.38 & 1.26 & \textbf{0.99} & \textbf{2.53} & \textbf{7.55}
                          & \textbf{2.09} & \textbf{1.28} & 3.57 & \textbf{0.81} & \textbf{0.16}
                          \\
 
       \bottomrule
     \end{tabular}
   }
 
   \caption{Quantitative results (second part) of registration on the Objaverse dataset. 
            $\Delta \mathbf{R}$ denotes the relative rotation errors 
            in degree, $\Delta \mathbf{t}$ denotes the relative translation errors multiplied by 1e2 with unknown scales. 
            FGR~\cite{DBLP:conf/eccv/ZhouPK16} denotes the fast global matching method, $\text{Ours}_{\text{df}}$ denotes our 
            method with surface fields replaced by density fields.} 
 
   \label{table:quantitive_objaverse_registration_part34}
\end{table*}


\begin{figure*}[h]
  \centering

  \begin{subfigure}[b]{1.0\textwidth}
    \includegraphics[width=0.95\linewidth]{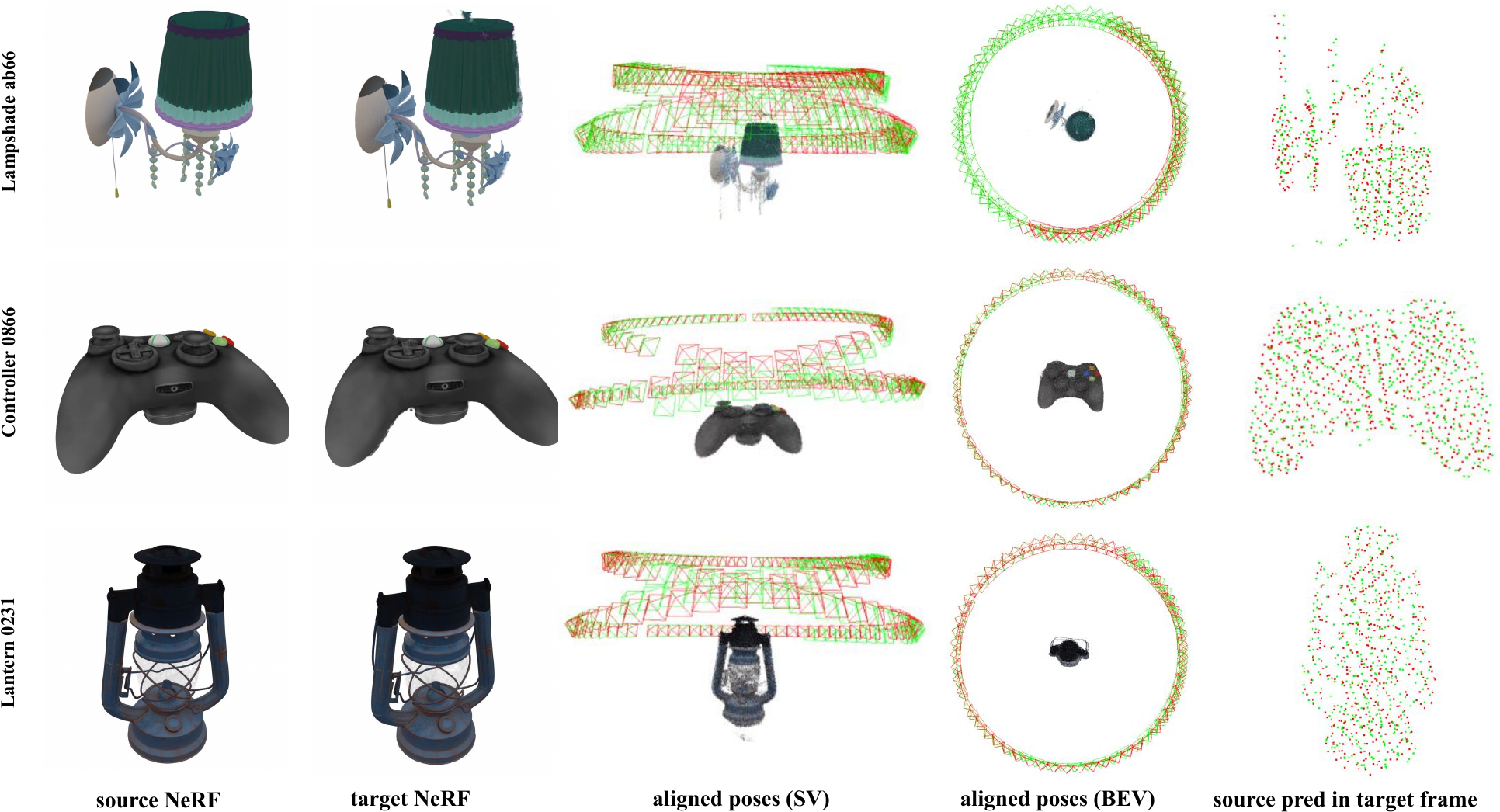}
    \label{fig:objaverse1}
  \end{subfigure}

  \caption{\textbf{The qualitative results on Objaverse~\cite{DBLP:journals/corr/abs-2212-08051} dataset after NeRF registration}. 
           From left to right are respectively the rendered images by the source NeRF model, the rendered images by the target NeRF 
           model, the side view (SV) of the aligned camera poses, the birds-eye-view (BEV) of the aligned camera poses, the
           concatenated predictions by transforming the source prediction to the target NeRF's coordinate frame. \textcolor{red}{red} 
           and \textcolor{green}{green}, respectively, denote the results from source NeRF and target NeRF.}
  \label{fig:objaverse_visualization}
\end{figure*}

\vspace{-3mm}
\paragraph{Evaluation.}
NeRF2NeRF~\cite{DBLP:journals/corr/abs-2211-01600} needs human annotated keypoints for initialization, which are not available on the dataset.
Therefore, we do not evaluate NeRF2NeRF on this dataset and use Fast Global Registration 
(FGR)~\cite{DBLP:conf/eccv/ZhouPK16} as the baseline. We also compared it against the state-of-the-art deep point cloud registration method 
REGTR~\cite{DBLP:conf/cvpr/YewL22}. For FGR and REGTR, we extract the voxel grid of each NeRF block to a point cloud and use the pairwise 
point clouds as input to them. For REGTR, we use the model that is pre-trained on the 3DMatch~\cite{DBLP:conf/cvpr/ZengSNFXF17} dataset 
provided by the author. We do not retrain REGTR on the Objaverse dataset since the ground-truth overlapping labels are not available on 
this dataset. We also evaluated our method $\text{Ours}_{\text{df}}$ with the surface fields replaced by the density fields as a comparison.

\begin{figure}[htbp]
   \centering
    \includegraphics[width=0.9\linewidth]{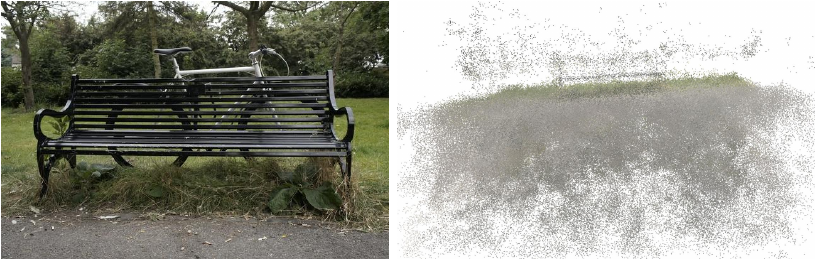}
   \caption{Our method failed on unbounded scenes where noisy points are extracted from the occupancy grid.}
   \vspace{-3mm}
   \label{fig:lego_failed}
 \end{figure}

\noindent \textbf{Results.}
The quantitative results can be seen from Tab.~\ref{table:quantitive_objaverse_registration_part12} and Tab.~\ref{table:quantitive_objaverse_registration_part34}. 
We report the relative rotation errors (RRE) $\Delta \mathbf{R}$ (in degree) and the 
relative translation errors (RTE) $\Delta \mathbf{t}$ as metrics. Note that the scale of translation is unknown and we multiply 
$\Delta \mathbf{t}$ by $1e2$. As we can see, FGR~\cite{DBLP:conf/eccv/ZhouPK16} failed in most of the scenes. We think it is the low resolution of 
our voxel grids that makes FGR~\cite{DBLP:conf/eccv/ZhouPK16} fail to find the correspondences. REGTR~\cite{DBLP:conf/cvpr/YewL22} also 
fails to find the correct transformations in almost all the scenes. It is worse than FGR in both rotations and translations. 
We also find 
very poor generalization ability of $\text{Ours}_{\text{df}}$. 
We conjecture that it is 
due to the density fields being too noisy and not unique for identifying per-scene geometry. In contrast, ``Ours'' achieves the best results 
among almost all the scenes -- since the network can be regularized to focus on the scene geometry properties by leveraging the surface fields.

\begin{figure}[h] 
  \centering

  \includegraphics[width=1.0\linewidth]{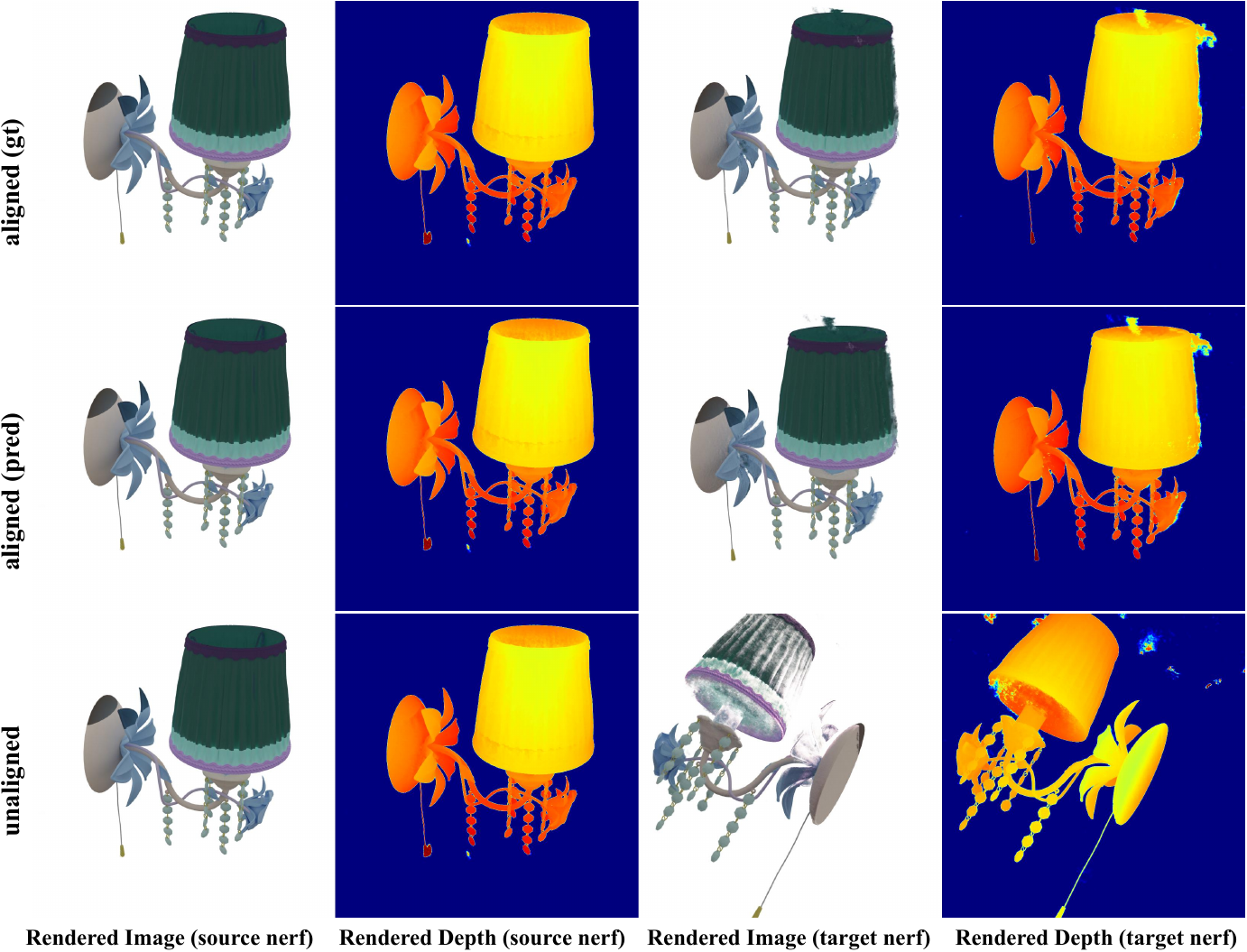}

  \caption{View synthesis comparison on object `Lampshade a166'. Top row: results from ground truth transformation. 
           Middle row: results from the predicted 
           transformation. Bottom row: results without applying any transformations.}
  \label{fig:vs_render_odeon_ab66}
  \vspace{-3mm}
\end{figure}

We present some qualitative results in Fig.~\ref{fig:objaverse_visualization}. To visualize the rendered images, we 
first transform the camera poses $\mathbf{P}_{\text{source}}$ in the source NeRF model to the target NeRF and obtain the transformed 
camera poses $\mathbf{P}_{\text{source}}^{'}$, and then we concatenate the transformed camera poses with the camera poses in target NeRF 
$\mathbf{P}_{\text{render}}=\text{concat}([\mathbf{P}_{\text{source}}^{'},\ \mathbf{P}_{\text{target}}])$.
We use $\mathbf{P}_{\text{source}}$ as the camera trajectories to synthesize images for the target NeRF model.
Similarly, we use $\mathbf{P}_{\text{render}}^{'}=\text{concat}([\mathbf{P}_{\text{source}},\ \mathbf{P}_{\text{target}}^{'}])$
for the source NeRF model to synthesize images, where $\mathbf{P}_{\text{target}}^{'}$ is the transformed camera poses from target NeRF to 
source NeRF. The results are respectively given in columns 1 and 2. We also visualize the aligned camera poses 
$\mathbf{P}_{\text{render}}$ in column 3 and 4. The red and green color respectively denotes the camera poses aligned by 
\textcolor{red}{ground truth} and our \textcolor{green}{estimated} transformations. Moreover, we transform the 
\textcolor{red}{source} prediction to the \textcolor{green}{target} coordinate frame and visualize the result in the last column.
We further use ground truth transformation to align the camera poses to obtain $\mathbf{P}_{\text{render}}^{\text{gt}}$, and simply concatenate source 
and target camera poses together to obtain $\mathbf{P}_{\text{render}}^{\text{no trans}}$. Subsequently, we provide the target NeRF respectively with 
$\mathbf{P}_{\text{render}}^{\text{gt}}, \mathbf{P}_{\text{render}}, \mathbf{P}_{\text{render}}^{\text{no trans}}$ to render RGB images 
and depth images. Some of the results are shown in Fig.~\ref{fig:vs_render_odeon_ab66}.
%

\vspace{-3mm}
\paragraph{Ablation Studies.} We further show the mean of RRE and RTE in Tab.~\ref{table:ablations} to ablate our method. 
``$\text{w.o.}\ \text{conf}$" denotes training our method without the confidence loss, ``$\text{w.o.}\ \text{sf}$" denotes 
training our network without the surface field loss, ``$\text{Ours}_{\text{df}}$" denotes our method with the surface fields replaced by 
the density fields. We can see that the surface fields are critical to our network training.
\begin{table}[htbp]
  \centering
  \resizebox{0.45\textwidth}{!}{
    \begin{tabular}{l r r r r r r}
      \toprule

      \multirow{1}{*}{}  &
      \multicolumn{1}{c}{$\text{FGR}$~\cite{DBLP:conf/eccv/ZhouPK16}} &
      \multicolumn{1}{c}{$\text{REGTR}$~\cite{DBLP:conf/cvpr/YewL22}} &
      \multicolumn{1}{c}{$\text{w.o.}\ \text{conf}$} &
      \multicolumn{1}{c}{$\text{w.o.}\ \text{sf}$} &  
      \multicolumn{1}{c}{$\text{Ours}_{\text{df}}$} &
      \multicolumn{1}{c}{Ours} \\

      \midrule

      $\Delta \mathbf{R}\ (^{\circ})$ & 61.59 & 113.78 & 71.84 & 101.17 & 86.22 & \textbf{9.67} \\
      
      \hline
      
      $\Delta \mathbf{t}\ (\times 1e2)$ & 13.50 & 43.31 & 12.97 & 20.35 & 16.06  & \textbf{3.85} \\

      \bottomrule
    \end{tabular}
  }
  \caption{Ablations studies of our method. The results are averaged on the 44 test objects.}
  \label{table:ablations}
  \vspace{-3mm}
\end{table}

\vspace{-3mm}
\paragraph{Performance Analysis.} To accelerate the network training, we do not query the NeRF model to obtain the voxel grid. Instead, 
we pre-compute the voxel grids $\mathbf{G}$ and the corresponding binary mask $\mathbf{M}$ for all NeRF blocks and store them on disks. 
The voxel grid $\mathbf{G}$ and binary mask $\mathbf{M}$ are loaded into memory at each iteration.  
During training, our network takes about $2.8$ seconds per iteration. 
The bottleneck on the training time 
is 
from loading $\mathbf{G}$, $\mathbf{M}$, and 
the source and target NeRF models. During inference, our model takes about $0.4$ seconds with the input voxel grid containing about 
$10$K points. Further acceleration can be achieved by pre-downsampling the voxel grid to a lower resolution.

\subsection{Further Discussion}
\paragraph{Limitations.} 
Our method has shown good performance in registering NeRF blocks. However, registering NeRF is still a challenging problem in large-scale scenes.
We summarize the limitations of our work as follows (more discussions are given in the supplementary):

\begin{itemize}

\item \textit{\textbf{Generalizability \vs out-of-distribution}} (OOD). While our method is generalizable to unseen 
\textbf{\textit{in-distribution}} scenes during testing, we postulate that performance would drop when tested on OOD scenes/object 
classes, \eg, training on indoor and testing on outdoor scenes, \etc.

\item \textit{\textbf{Application to real-world data $\&$ unbounded scenes}}. We emphasize that our training data contains real-world 
objects (\eg Shoes in Fig.~\ref{fig:training_data_overview}). Our method currently cannot be applied to unbounded scenes since NeRF is not good at 
geometry estimation. Consequently, incorrect geometries like floaters can influence the performance of our model. 
It means that our method can fail if the extracted occupancy grid contains too many noisy points (See Fig.~\ref{fig:lego_failed}). 
We argue that better results can be obtained by applying RANSAC~\cite{DBLP:journals/cacm/FischlerB81} to filter outliers based on our predicted correspondences, or training 
better NeRF blocks by utilizing depth supervision~\cite{DBLP:conf/cvpr/DengLZR22,DBLP:conf/cvpr/RoessleBMSN22}, or utilizing a robust 
loss~\cite{DBLP:journals/corr/abs-2302-00833} to ignoring floaters during training NeRF blocks.
In addition, for real-world data that contain background, techniques like~\cite{DBLP:journals/corr/abs-2304-12308} 
can also be applied to our method to get the interested objects. We leave this as our future work.

\item \textit{\textbf{Scale in the relative transformation.}} We follow the assumption of NeRF2NeRF that the scales for two NeRFs are 
the same, which can be violated in real-world settings.
Nonetheless, additional sensors such as IMU, wheel encoders, \etc, are easily available to get the absolute scale. For settings 
where only RGB images are available, the scale can be a problem. Additional scene priors are needed to fix the scale for RGB 
images as input.

\end{itemize}

\begin{figure}[H]
  \centering
  \includegraphics[width=1.0\linewidth]{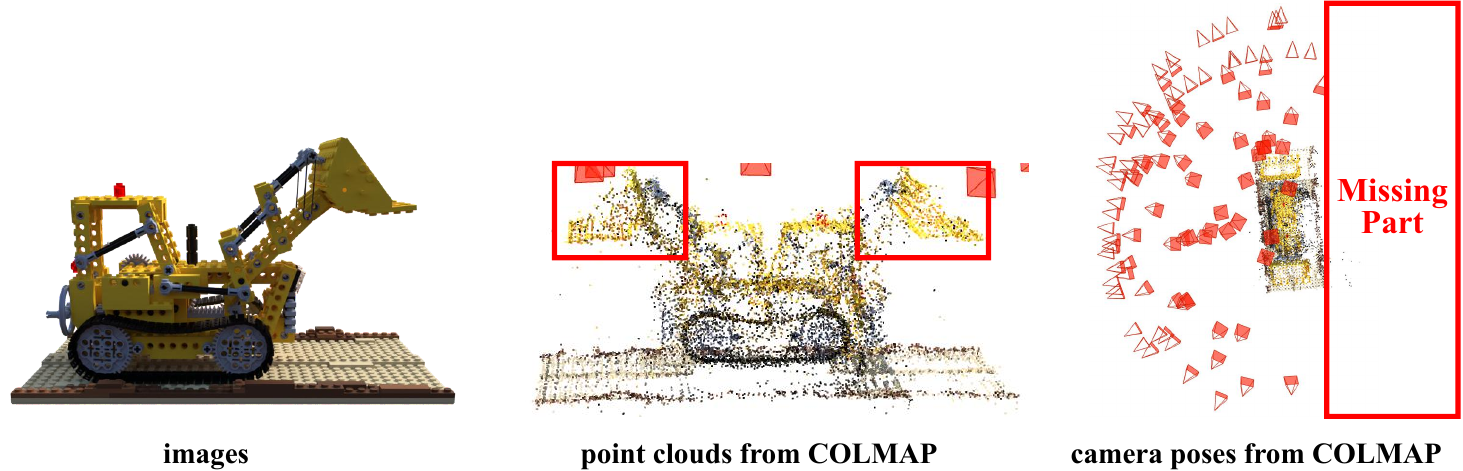}
  \caption{COLMAP failed on synthetic dataset due to wrong correspondences.}
  \label{fig:colmap_failed}
\end{figure}

\vspace{-4mm}
\paragraph{Why localization methods based on SfM tools are not compared?} 
A simple solution is to first synthesize images using NeRF. 
We can then use SfM to get 2D-2D correspondences from keypoints matcher and do triangulation to recover the 3D scene points. Consequently, localization-based methods such as perspective-n-points (PnP)~\cite{DBLP:conf/cvpr/KneipSS11} or iterative closest point (ICP)~\cite{DBLP:journals/pami/BeslM92,DBLP:journals/ivc/ChenM92} can be applied on the 3D scene points to register the NeRF models.
However, we argue that SfM is fragile in scenes where keypoint correspondences are difficult to establish. 
One failure case is given in Fig.~\ref{fig:colmap_failed}. As a result, all methods that rely on keypoint correspondences can potentially fail due to wrong matches. Particularly, it is often hard to obtain enough matches for texture-less scenes/objects. 
False matches can also occur due to changes in image appearance. We circumvent this problem by learning the correspondences from NeRF 
representations, \ie the density field, which is shown robust to image appearance changes in the experiments. Moreover, our method is an 
end-to-end solution and therefore can be much faster than other methods that rely on keypoints, \eg, iNeRF~\cite{DBLP:conf/iros/LinFBRIL21} 
takes more than 50 secs to register an image in an existing NeRF model (\emph{c.f.} Fig. 4 of the iNeRF paper), while ours only takes 0.4 
secs to register two NeRFs.
%
%
%
\vspace{-5mm}
\paragraph{Why not register images in the same coordinate frame by global bundle adjustment (BA)?}
We argue that there are cases where using BA to recover all poses may not be the best option:
\begin{itemize}
\item \textbf{Robustness}. BA relies on good keypoint correspondences which can be 
challenging to obtain in texture-less scenes, \etc. In contrast,
our DReg-NeRF leverages NeRF features for registration without explicit correspondence search on the images.
\item \textbf{Scalability and efficiency}. Images of a large scene can be collected in smaller batches. It is more scalable and efficient
to build smaller NeRF models on each batch of images and subsequently do NeRF registration to get the NeRF model of the full scene.
\item \textbf{Modularity}. It is easier to update a modular NeRF model. Any module can be easily replaced or added via NeRF registration.

\end{itemize}

\section{Conclusion}

In conclusion, we have proposed a novel network architecture that registers NeRF blocks into the same coordinate frame. Unlike existing methods, 
our method does not rely on any initialization and human-annotated keypoints. We constructed a dataset with $1,700+$ objects where 
images are rendered from 3D textured meshes of the Objaverse dataset. We train our method on our constructed dataset. Our method surpasses 
the state-of-the-art traditional and learning-based point cloud registration methods when evaluated on the test set.

\paragraph{Acknowledgement.} 
This research work is supported by the Agency for Science, Technology and Research (A*STAR) under its MTC Programmatic Funds (Grant No. M23L7b0021). 

{\small
\bibliographystyle{ieee_fullname}
\bibliography{egbib}

\begin{thebibliography}{10}\itemsep=-1pt

\bibitem{DBLP:conf/cvpr/Barron19}
Jonathan~T. Barron.
\newblock A general and adaptive robust loss function.
\newblock In {\em {IEEE} Conference on Computer Vision and Pattern
  Recognition}, pages 4331--4339, 2019.

\bibitem{DBLP:journals/pami/BeslM92}
Paul~J. Besl and Neil~D. McKay.
\newblock A method for registration of 3-d shapes.
\newblock {\em {IEEE} Trans. Pattern Anal. Mach. Intell.}, 14(2):239--256,
  1992.

\bibitem{DBLP:journals/corr/abs-2304-12308}
Jiazhong Cen, Zanwei Zhou, Jiemin Fang, Wei Shen, Lingxi Xie, Dongsheng Jiang,
  Xiaopeng Zhang, and Qi Tian.
\newblock Segment anything in 3d with nerfs.
\newblock {\em CoRR}, abs/2304.12308, 2023.

\bibitem{DBLP:conf/eccv/ChenXGYS22}
Anpei Chen, Zexiang Xu, Andreas Geiger, Jingyi Yu, and Hao Su.
\newblock Tensorf: Tensorial radiance fields.
\newblock In {\em Computer Vision - {ECCV} 2022 - 17th European Conference},
  volume 13692, pages 333--350, 2022.

\bibitem{chen2023dbarf}
Yu Chen and Gim~Hee Lee.
\newblock Dbarf: Deep bundle-adjusting generalizable neural radiance fields.
\newblock In {\em Proceedings of the IEEE/CVF Conference on Computer Vision and
  Pattern Recognition}, pages 24--34, 2023.

\bibitem{DBLP:journals/ivc/ChenM92}
Yang Chen and G{\'{e}}rard~G. Medioni.
\newblock Object modelling by registration of multiple range images.
\newblock {\em Image Vis. Comput.}, 10(3):145--155, 1992.

\bibitem{DBLP:journals/pr/ChenSCW20}
Yu Chen, Shuhan Shen, Yisong Chen, and Guoping Wang.
\newblock Graph-based parallel large scale structure from motion.
\newblock {\em Pattern Recognition.}, 107:107537, 2020.

\bibitem{DBLP:journals/corr/abs-2301-12135}
Yu Chen, Zihao Yu, Shu Song, Tianning Yu, Jianming Li, and Gim~Hee Lee.
\newblock Adasfm: From coarse global to fine incremental adaptive structure
  from motion.
\newblock In {\em 2023 IEEE International Conference on Robotics and Automation
  (ICRA)}, pages 2054--2061, 2023.

\bibitem{DBLP:conf/cvpr/Chen0K21}
Yu Chen, Ji Zhao, and Laurent Kneip.
\newblock Hybrid rotation averaging: {A} fast and robust rotation averaging
  approach.
\newblock In {\em {IEEE} Conference on Computer Vision and Pattern
  Recognition}, pages 10358--10367, 2021.

\bibitem{DBLP:journals/corr/abs-2212-08051}
Matt Deitke, Dustin Schwenk, Jordi Salvador, Luca Weihs, Oscar Michel, Eli
  VanderBilt, Ludwig Schmidt, Kiana Ehsani, Aniruddha Kembhavi, and Ali
  Farhadi.
\newblock Objaverse: {A} universe of annotated 3d objects.
\newblock {\em CoRR}, abs/2212.08051, 2022.

\bibitem{DBLP:conf/cvpr/DengLZR22}
Kangle Deng, Andrew Liu, Jun{-}Yan Zhu, and Deva Ramanan.
\newblock Depth-supervised nerf: Fewer views and faster training for free.
\newblock In {\em {IEEE/CVF} Conference on Computer Vision and Pattern
  Recognition}, pages 12872--12881, 2022.

\bibitem{DBLP:journals/cacm/FischlerB81}
Martin~A. Fischler and Robert~C. Bolles.
\newblock Random sample consensus: {A} paradigm for model fitting with
  applications to image analysis and automated cartography.
\newblock {\em Commun. {ACM}}, 24(6):381--395, 1981.

\bibitem{DBLP:conf/cvpr/Fridovich-KeilY22}
Sara Fridovich{-}Keil, Alex Yu, Matthew Tancik, Qinhong Chen, Benjamin Recht,
  and Angjoo Kanazawa.
\newblock Plenoxels: Radiance fields without neural networks.
\newblock In {\em {IEEE/CVF} Conference on Computer Vision and Pattern
  Recognition}, pages 5491--5500, 2022.

\bibitem{DBLP:journals/corr/abs-2211-01600}
Lily Goli, Daniel Rebain, Sara Sabour, Animesh Garg, and Andrea Tagliasacchi.
\newblock nerf2nerf: Pairwise registration of neural radiance fields.
\newblock {\em CoRR}, abs/2211.01600, 2022.

\bibitem{DBLP:conf/cvpr/HeZRS16}
Kaiming He, Xiangyu Zhang, Shaoqing Ren, and Jian Sun.
\newblock Deep residual learning for image recognition.
\newblock In {\em 2016 {IEEE} Conference on Computer Vision and Pattern
  Recognition}, pages 770--778, 2016.

\bibitem{DBLP:conf/cvpr/HuangGUWS21}
Shengyu Huang, Zan Gojcic, Mikhail Usvyatsov, Andreas Wieser, and Konrad
  Schindler.
\newblock Predator: Registration of 3d point clouds with low overlap.
\newblock In {\em {IEEE} Conference on Computer Vision and Pattern
  Recognition}, pages 4267--4276, 2021.

\bibitem{DBLP:journals/corr/KingmaB14}
Diederik~P. Kingma and Jimmy Ba.
\newblock Adam: {A} method for stochastic optimization.
\newblock In {\em 3rd International Conference on Learning Representations},
  2015.

\bibitem{DBLP:conf/cvpr/KneipSS11}
Laurent Kneip, Davide Scaramuzza, and Roland Siegwart.
\newblock A novel parametrization of the perspective-three-point problem for a
  direct computation of absolute camera position and orientation.
\newblock In {\em {IEEE} Conference on Computer Vision and Pattern
  Recognition}, pages 2969--2976, 2011.

\bibitem{DBLP:journals/corr/abs-2210-04847}
Ruilong Li, Matthew Tancik, and Angjoo Kanazawa.
\newblock Nerfacc: {A} general nerf acceleration toolbox.
\newblock {\em CoRR}, abs/2210.04847, 2022.

\bibitem{DBLP:journals/access/LiWZ22a}
Tianyang Li, Jian Wang, and Tibing Zhang.
\newblock {L-DETR:} {A} light-weight detector for end-to-end object detection
  with transformers.
\newblock {\em {IEEE} Access}, 10:105685--105692, 2022.

\bibitem{DBLP:conf/iccv/LinM0L21}
Chen{-}Hsuan Lin, Wei{-}Chiu Ma, Antonio Torralba, and Simon Lucey.
\newblock {BARF:} bundle-adjusting neural radiance fields.
\newblock In {\em 2021 {IEEE/CVF} International Conference on Computer Vision},
  pages 5721--5731, 2021.

\bibitem{DBLP:conf/cvpr/LinDGHHB17}
Tsung{-}Yi Lin, Piotr Doll{\'{a}}r, Ross~B. Girshick, Kaiming He, Bharath
  Hariharan, and Serge~J. Belongie.
\newblock Feature pyramid networks for object detection.
\newblock In {\em 2017 {IEEE} Conference on Computer Vision and Pattern
  Recognition}, pages 936--944, 2017.

\bibitem{DBLP:conf/iros/LinFBRIL21}
Yen{-}Chen Lin, Pete Florence, Jonathan~T. Barron, Alberto Rodriguez, Phillip
  Isola, and Tsung{-}Yi Lin.
\newblock inerf: Inverting neural radiance fields for pose estimation.
\newblock In {\em {IEEE/RSJ} International Conference on Intelligent Robots and
  Systems}, pages 1323--1330, 2021.

\bibitem{DBLP:conf/iccv/LindenbergerSLP21}
Philipp Lindenberger, Paul{-}Edouard Sarlin, Viktor Larsson, and Marc
  Pollefeys.
\newblock Pixel-perfect structure-from-motion with featuremetric refinement.
\newblock In {\em 2021 {IEEE/CVF} International Conference on Computer Vision},
  pages 5967--5977, 2021.

\bibitem{DBLP:conf/nips/LiuGLCT20}
Lingjie Liu, Jiatao Gu, Kyaw~Zaw Lin, Tat{-}Seng Chua, and Christian Theobalt.
\newblock Neural sparse voxel fields.
\newblock In {\em Advances in Neural Information Processing Systems 33}, 2020.

\bibitem{DBLP:conf/iclr/LoshchilovH19}
Ilya Loshchilov and Frank Hutter.
\newblock Decoupled weight decay regularization.
\newblock In {\em 7th International Conference on Learning Representations},
  2019.

\bibitem{DBLP:conf/eccv/MildenhallSTBRN20}
Ben Mildenhall, Pratul~P. Srinivasan, Matthew Tancik, Jonathan~T. Barron, Ravi
  Ramamoorthi, and Ren Ng.
\newblock Nerf: Representing scenes as neural radiance fields for view
  synthesis.
\newblock In {\em Computer Vision - {ECCV} 2020 - 16th European Conference},
  volume 12346, pages 405--421, 2020.

\bibitem{DBLP:journals/tog/MullerESK22}
Thomas M{\"{u}}ller, Alex Evans, Christoph Schied, and Alexander Keller.
\newblock Instant neural graphics primitives with a multiresolution hash
  encoding.
\newblock {\em {ACM} Trans. Graph.}, 41(4):102:1--102:15, 2022.

\bibitem{DBLP:journals/corr/abs-2211-12544}
Casey Peat, Oliver Batchelor, Richard~D. Green, and James Atlas.
\newblock Zero nerf: Registration with zero overlap.
\newblock {\em CoRR}, abs/2211.12544, 2022.

\bibitem{DBLP:conf/cvpr/RoessleBMSN22}
Barbara Roessle, Jonathan~T. Barron, Ben Mildenhall, Pratul~P. Srinivasan, and
  Matthias Nie{\ss}ner.
\newblock Dense depth priors for neural radiance fields from sparse input
  views.
\newblock In {\em {IEEE/CVF} Conference on Computer Vision and Pattern
  Recognition}, pages 12882--12891, 2022.

\bibitem{DBLP:conf/icra/RusuBB09}
Radu~Bogdan Rusu, Nico Blodow, and Michael Beetz.
\newblock Fast point feature histograms {(FPFH)} for 3d registration.
\newblock In {\em 2009 {IEEE} International Conference on Robotics and
  Automation}, pages 3212--3217, 2009.

\bibitem{DBLP:journals/corr/abs-2302-00833}
Sara Sabour, Suhani Vora, Daniel Duckworth, Ivan Krasin, David~J. Fleet, and
  Andrea Tagliasacchi.
\newblock Robustnerf: Ignoring distractors with robust losses.
\newblock {\em CoRR}, abs/2302.00833, 2023.

\bibitem{DBLP:conf/cvpr/SarlinDMR20}
Paul{-}Edouard Sarlin, Daniel DeTone, Tomasz Malisiewicz, and Andrew
  Rabinovich.
\newblock Superglue: Learning feature matching with graph neural networks.
\newblock In {\em 2020 {IEEE/CVF} Conference on Computer Vision and Pattern
  Recognition}, pages 4937--4946, 2020.

\bibitem{DBLP:journals/corr/abs-2209-02417}
Andrea Tagliasacchi and Ben Mildenhall.
\newblock Volume rendering digest (for nerf).
\newblock {\em CoRR}, abs/2209.02417, 2022.

\bibitem{DBLP:conf/cvpr/TancikCYPMSBK22}
Matthew Tancik, Vincent Casser, Xinchen Yan, Sabeek Pradhan, Ben~P. Mildenhall,
  Pratul~P. Srinivasan, Jonathan~T. Barron, and Henrik Kretzschmar.
\newblock Block-nerf: Scalable large scene neural view synthesis.
\newblock In {\em {IEEE/CVF} Conference on Computer Vision and Pattern
  Recognition}, pages 8238--8248, 2022.

\bibitem{DBLP:conf/3dim/ThomasGDM18}
Hugues Thomas, Fran{\c{c}}ois Goulette, Jean{-}Emmanuel Deschaud, and Beatriz
  Marcotegui.
\newblock Semantic classification of 3d point clouds with multiscale spherical
  neighborhoods.
\newblock In {\em 2018 International Conference on 3D Vision, 3DV 2018}, pages
  390--398, 2018.

\bibitem{DBLP:conf/iccv/ThomasQDMGG19}
Hugues Thomas, Charles~R. Qi, Jean{-}Emmanuel Deschaud, Beatriz Marcotegui,
  Fran{\c{c}}ois Goulette, and Leonidas~J. Guibas.
\newblock Kpconv: Flexible and deformable convolution for point clouds.
\newblock In {\em 2019 {IEEE/CVF} International Conference on Computer Vision},
  pages 6410--6419, 2019.

\bibitem{DBLP:conf/cvpr/TurkiRS22}
Haithem Turki, Deva Ramanan, and Mahadev Satyanarayanan.
\newblock Mega-nerf: Scalable construction of large-scale nerfs for virtual
  fly- throughs.
\newblock In {\em {IEEE/CVF} Conference on Computer Vision and Pattern
  Recognition}, pages 12912--12921, 2022.

\bibitem{DBLP:journals/pami/Umeyama91}
Shinji Umeyama.
\newblock Least-squares estimation of transformation parameters between two
  point patterns.
\newblock {\em {IEEE} Trans. Pattern Anal. Mach. Intell.}, 13(4):376--380,
  1991.

\bibitem{DBLP:journals/corr/abs-1807-03748}
A{\"{a}}ron van~den Oord, Yazhe Li, and Oriol Vinyals.
\newblock Representation learning with contrastive predictive coding.
\newblock {\em CoRR}, abs/1807.03748, 2018.

\bibitem{DBLP:conf/nips/VaswaniSPUJGKP17}
Ashish Vaswani, Noam Shazeer, Niki Parmar, Jakob Uszkoreit, Llion Jones,
  Aidan~N. Gomez, Lukasz Kaiser, and Illia Polosukhin.
\newblock Attention is all you need.
\newblock In {\em Advances in Neural Information Processing Systems}, pages
  5998--6008, 2017.

\bibitem{DBLP:conf/iccv/WangS19}
Yue Wang and Justin Solomon.
\newblock Deep closest point: Learning representations for point cloud
  registration.
\newblock In {\em 2019 {IEEE/CVF} International Conference on Computer Vision},
  pages 3522--3531, 2019.

\bibitem{DBLP:conf/cvpr/YewL22}
Zi~Jian Yew and Gim~Hee Lee.
\newblock {REGTR:} end-to-end point cloud correspondences with transformers.
\newblock In {\em {IEEE/CVF} Conference on Computer Vision and Pattern
  Recognition}, pages 6667--6676, 2022.

\bibitem{DBLP:conf/iccv/YuLT0NK21}
Alex Yu, Ruilong Li, Matthew Tancik, Hao Li, Ren Ng, and Angjoo Kanazawa.
\newblock Plenoctrees for real-time rendering of neural radiance fields.
\newblock In {\em 2021 {IEEE/CVF} International Conference on Computer Vision},
  pages 5732--5741, 2021.

\bibitem{DBLP:conf/cvpr/ZengSNFXF17}
Andy Zeng, Shuran Song, Matthias Nie{\ss}ner, Matthew Fisher, Jianxiong Xiao,
  and Thomas~A. Funkhouser.
\newblock 3dmatch: Learning local geometric descriptors from {RGB-D}
  reconstructions.
\newblock In {\em 2017 {IEEE} Conference on Computer Vision and Pattern
  Recognition}, pages 199--208, 2017.

\bibitem{DBLP:conf/eccv/ZhouPK16}
Qian{-}Yi Zhou, Jaesik Park, and Vladlen Koltun.
\newblock Fast global registration.
\newblock In Bastian Leibe, Jiri Matas, Nicu Sebe, and Max Welling, editors,
  {\em Computer Vision - {ECCV} 2016 - 14th European Conference}, volume 9906
  of {\em Lecture Notes in Computer Science}, pages 766--782.

\end{thebibliography}
}

\newpage
\clearpage
\section{APPENDIX}

\subsection{NeRF Network Architecture}

\begin{figure*}[h]
   \centering
 
   \includegraphics[width=0.99\linewidth]{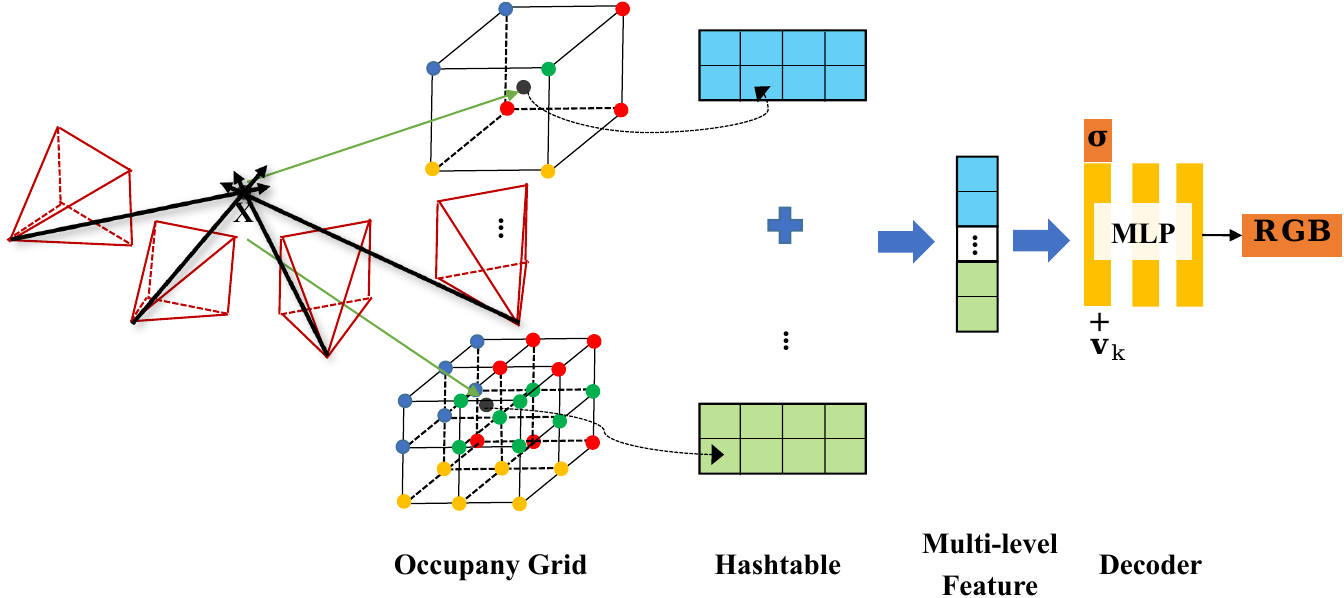}
 
   \caption{\textbf{Network architecture of our used NeRF}. A single-resolution occupancy grid is used to skip empty space.
      The dimension of each hidden layer is 64. The view direction is concatenated with the feature embedding after the first 
      hidden layer without applying positional encodings.
   }
 
   \label{fig:nerf_architecture}
 \end{figure*}

We present the network architecture of our used NeRF network in Fig.~\ref{fig:nerf_architecture}. The resolution level is 16.
The number of hash table entries in each level is $2^{19}$, where the feature dimension of each hash table entry is 2. The coarsest level 
is 16. We use NeRFAcc~\cite{DBLP:journals/corr/abs-2210-04847} to train NeRF models, where only a single resolution occupancy grid is used 
to skip empty space instead of multi-resolution occupancy grids as in the original InstantNGP implementation.

\subsection{More Qualitative Results}

\begin{figure*}[h]
   \centering
 
   \begin{subfigure}[b]{1.0\textwidth}
     \includegraphics[width=0.95\linewidth]{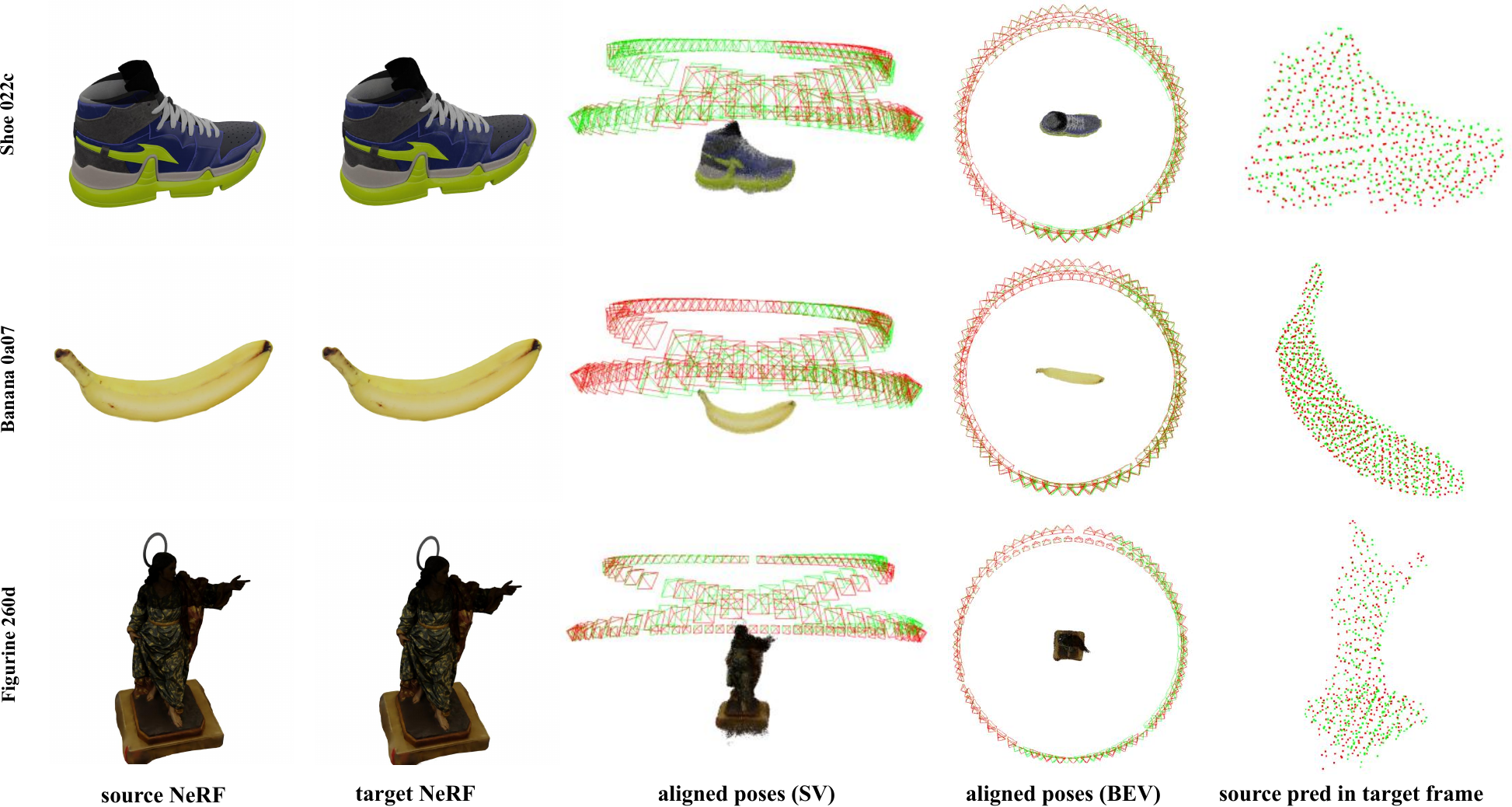}
     \label{fig:objaverse2}
   \end{subfigure}

   \caption{\textbf{The qualitative results on Objaverse~\cite{DBLP:journals/corr/abs-2212-08051} dataset after NeRF registration}. 
            From left to right are respectively the rendered images by the source NeRF model, the rendered images by the target NeRF 
            model, the side view (SV) of the aligned camera poses, the birds-eye-view (BEV) of the aligned camera poses, the
            concatenated predictions by transforming the source prediction to the target NeRF's coordinate frame. \textcolor{red}{red} 
            and \textcolor{green}{green}, respectively, denote the results from source NeRF and target NeRF.}
 
   \label{fig:objaverse_visualization2}
 \end{figure*}

\begin{figure*}[htbp]
  \centering

   \includegraphics[width=1.0\linewidth]{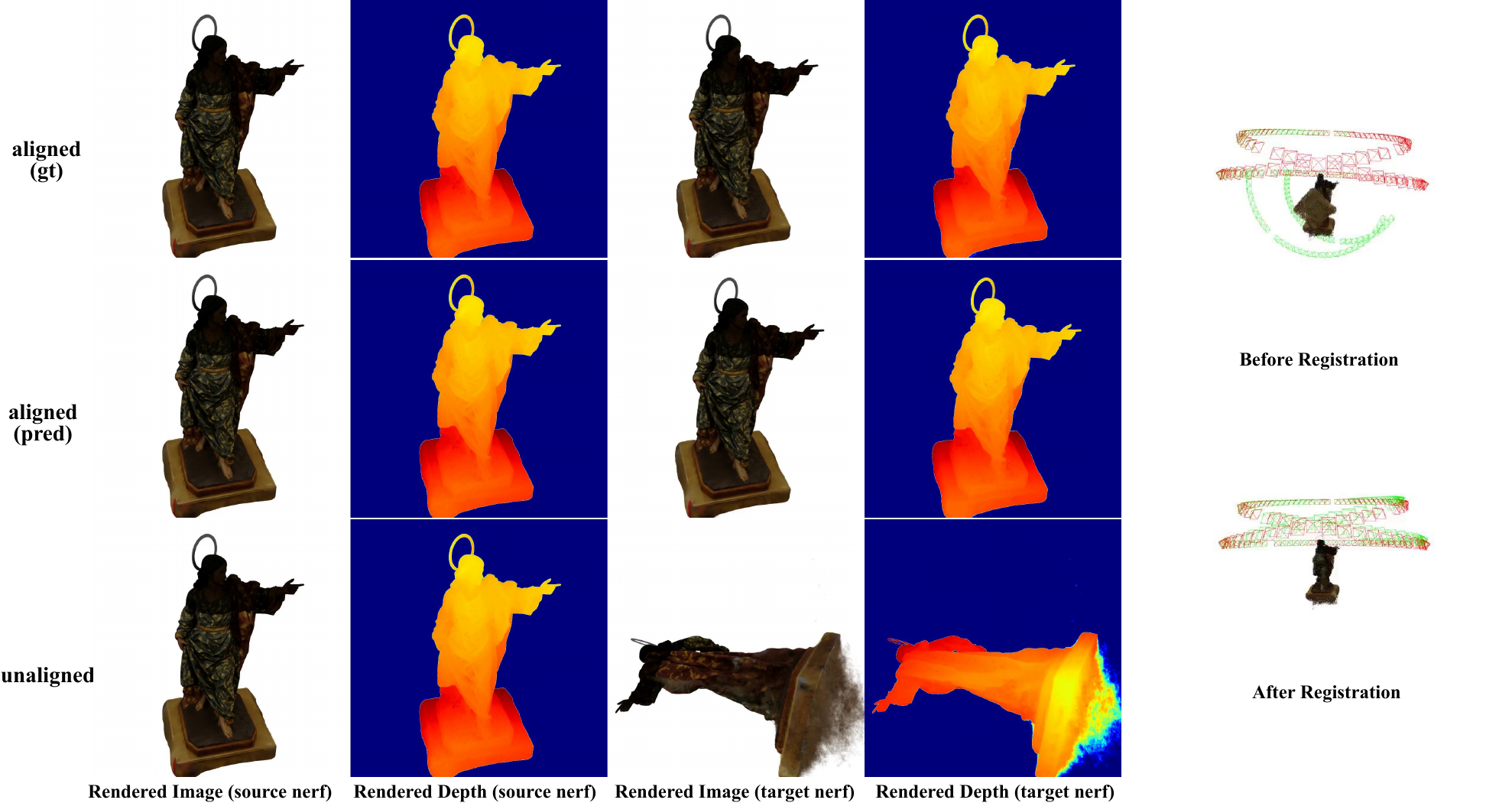}
   \includegraphics[width=1.0\linewidth]{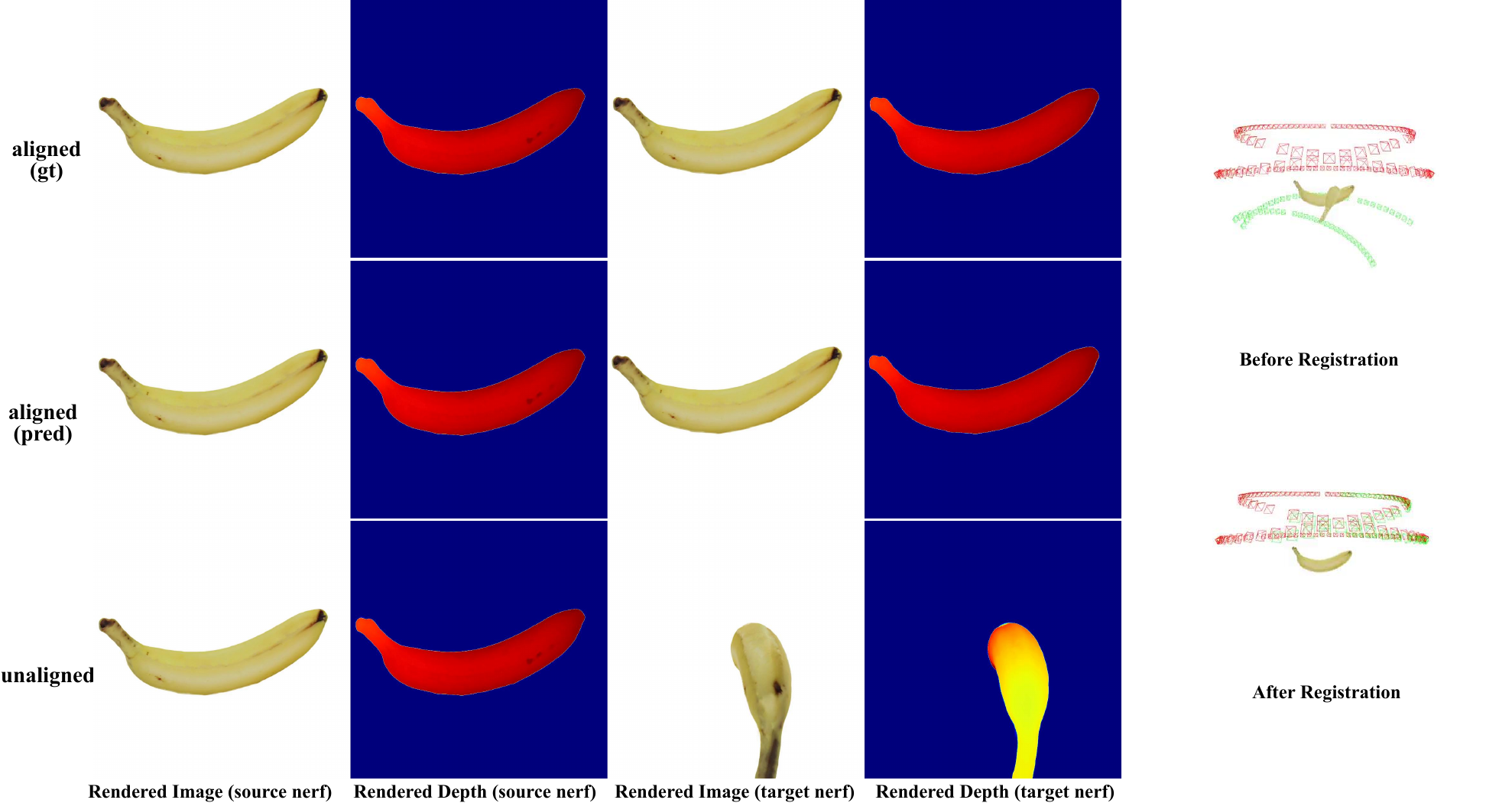}

  \caption{\textbf{View synthesis comparison.}
           Left: (1) Top row: results from ground truth transformation; (2) Middle row: results from the predicted transformation; 
           (3) Bottom row: results without applying the transformation.
           Right: (1) Camera poses and occupancy grids before registration; (2) Camera poses and occupancy grids after registration.
   }
  \label{fig:vs_render1}
\end{figure*}

\begin{figure*}[htbp]
   \centering
 
   \includegraphics[width=1.0\linewidth]{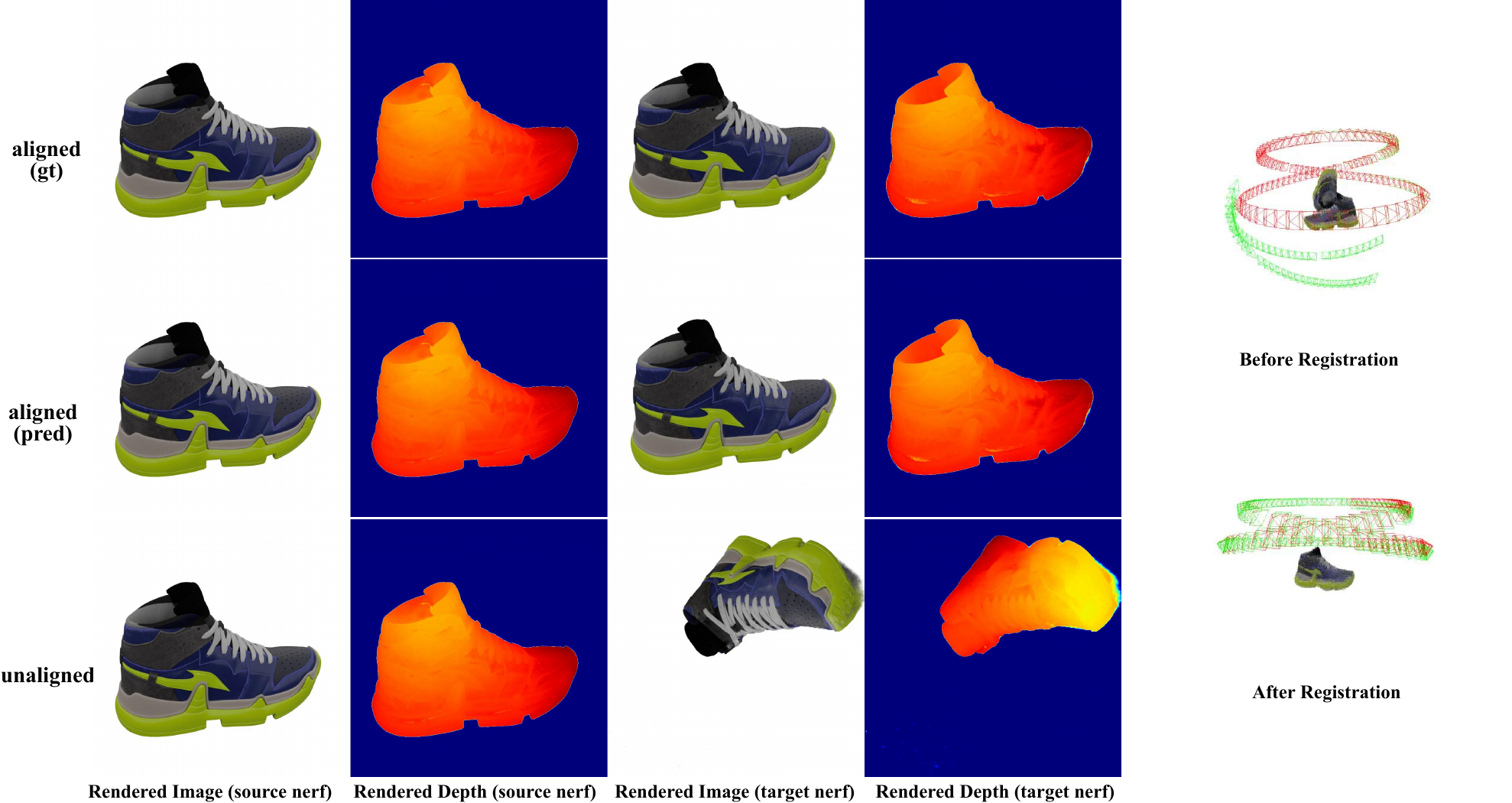}
   \includegraphics[width=1.0\linewidth]{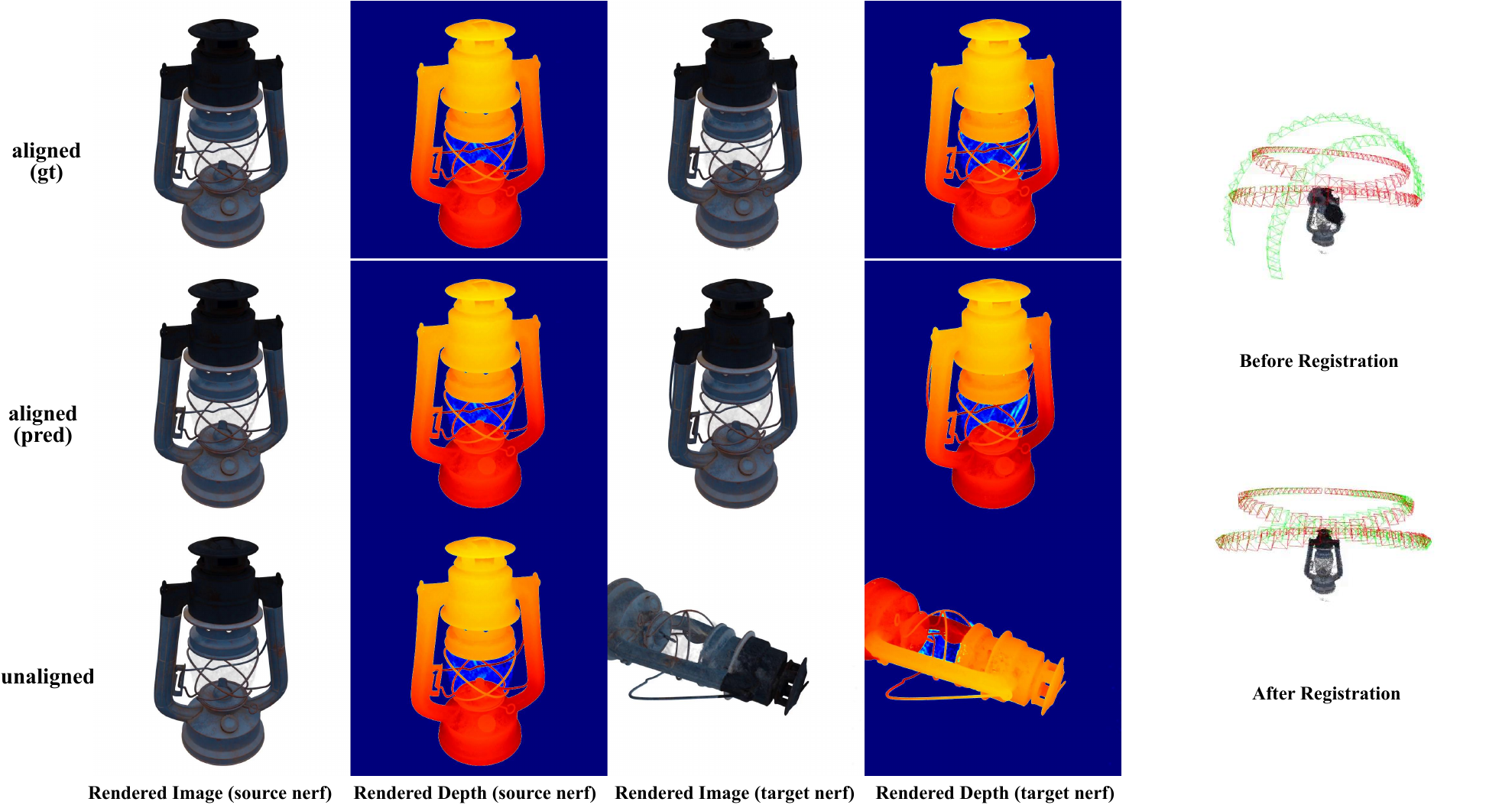}
 
   \caption{\textbf{View synthesis comparison on object.}
            Left: (1) Top row: results from ground truth transformation; (2) Middle row: results from the predicted transformation; 
            (3) Bottom row: results without applying the transformation.
            Right: (1) Camera poses and occupancy grids before registration; (2) Camera poses and occupancy grids after registration.
   }
   \label{fig:vs_render2}
\end{figure*}

\begin{figure*}[htbp]
   \centering
 
   \includegraphics[width=1.0\linewidth]{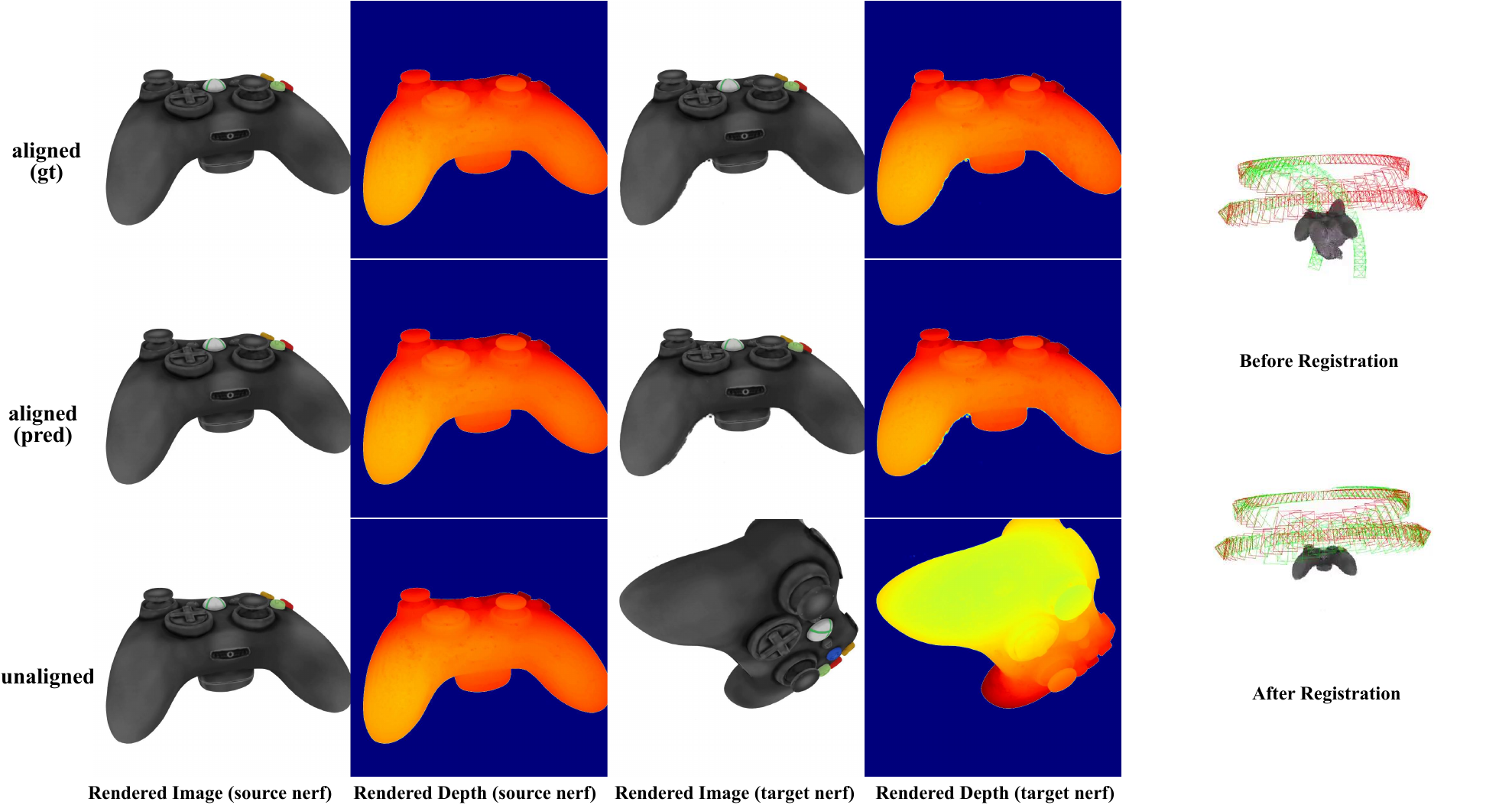}
   \includegraphics[width=1.0\linewidth]{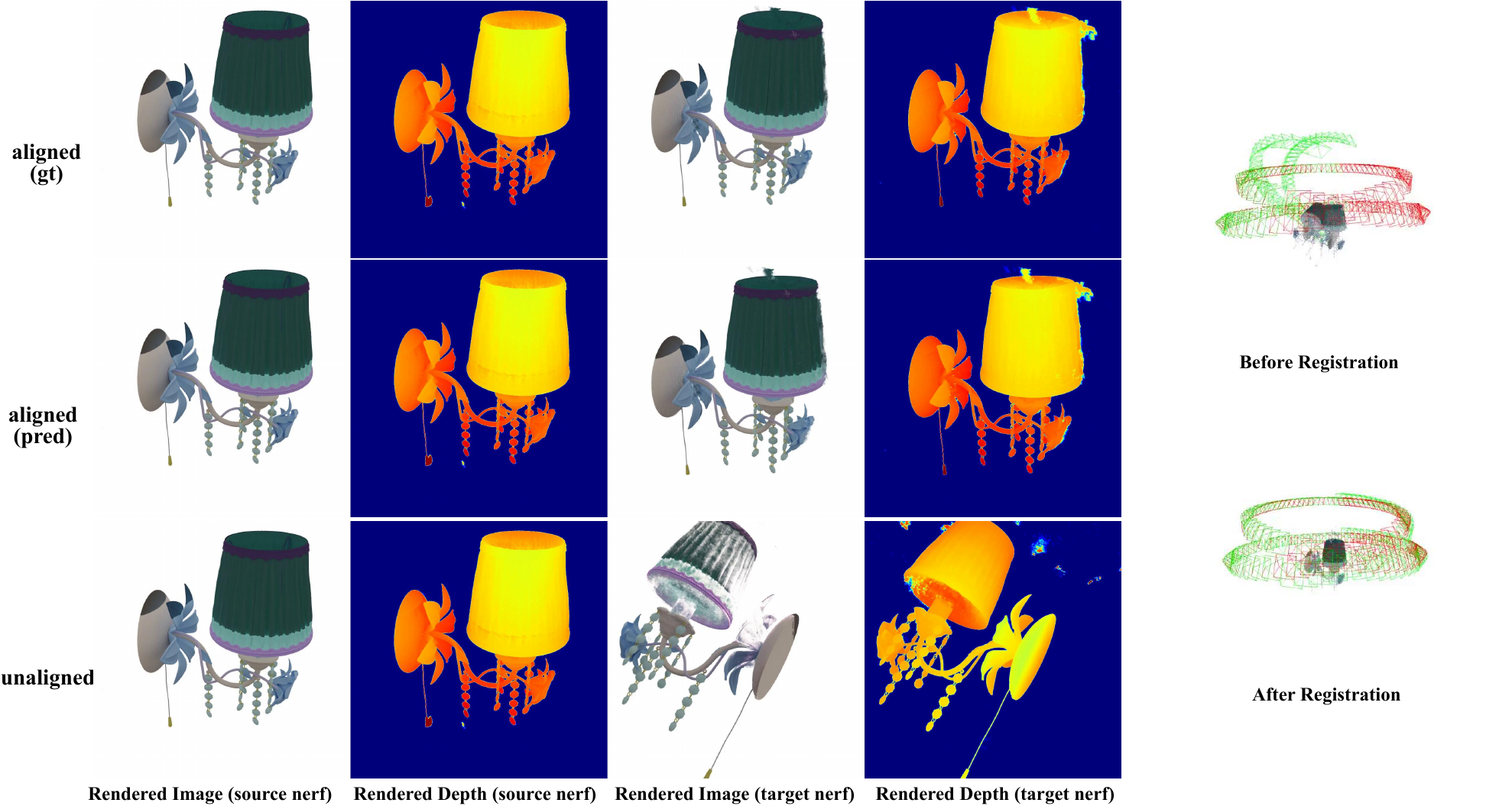}
 
   \caption{\textbf{View synthesis comparison.}
            Left: (1) Top row: results from ground truth transformation; (2) Middle row: results from the predicted transformation; 
            (3) Bottom row: results without applying the transformation.
            Right: (1) Camera poses and occupancy grids before registration; (2) Camera poses and occupancy grids after registration.
   }
   \label{fig:vs_render3}
 \end{figure*}

We present more qualitative results in Fig.~\ref{fig:objaverse_visualization2}. We further visualize the rendered RGB images and depth images 
in Fig.~\ref{fig:vs_render1}, Fig.~\ref{fig:vs_render2} and Fig.~\ref{fig:vs_render3}.
In the left part of each figure, we visualize the rendered RGB images and depth images, where the top row shows results from the ground truth 
transformation, the middle row shows results from the predicted transformation, and the bottom row shows results without applying 
transformation. The right part of each figure presents camera poses and occupancy grids before registration on the top row,
and the camera poses and occupancy grids after registration on the bottom row.


\end{document}